\pdfoutput=1
\documentclass[11pt]{article}

\usepackage{ACL2023}
\usepackage{times}
\usepackage{latexsym}

\usepackage[T1]{fontenc}
\usepackage[utf8]{inputenc}

\usepackage{microtype}

\usepackage{inconsolata}

\usepackage{times}
\usepackage{latexsym}
\usepackage{todonotes}
\usepackage{graphicx}
\usepackage{pifont}
\usepackage{siunitx}
\usepackage{tikz}
\usepackage{adjustbox}
\usepackage{subcaption}
\usepackage{xcolor}
\usepackage{amssymb}
\usepackage{amsmath}
\usepackage[nameinlink,noabbrev,capitalise,sort]{cleveref}
\usepackage{multirow}
\usepackage{soul}
\usepackage[most]{tcolorbox}
\tcbset{on line, 
        boxsep=4pt, left=0pt,right=0pt,top=0pt,bottom=0pt,
        colframe=white,  
        highlight math style={enhanced}
        }

\usepackage[T1]{fontenc}
\usepackage[utf8]{inputenc}

\usepackage{microtype}

\usepackage{inconsolata}
\usepackage{tabularx}
\usepackage{booktabs}       % professional-quality tables
\usepackage{enumitem}

\newcommand{\cmark}{\ding{51}}%
\newcommand{\xmark}{\ding{55}}%

\definecolor{darkgreen}{rgb}{1,50,32}
\definecolor{mygreen}{rgb}{0.0, 0.5, 0.0}
\definecolor{myred}{rgb}{0.8, 0.0, 0.0}
\definecolor{snowflake}{rgb}{0.26,0.45,0.72}
\definecolor{lightmagenta}{rgb}{0.95, 0.8, 1.0}
\usepackage{fontawesome}
\usepackage{tikz}
\usepackage{bbding} 
\usepackage[weather]{ifsym}

\title{Shaking Up VLMs: Comparing Transformers and Structured State Space Models for Vision \& Language Modeling}

\author{
    Georgios Pantazopoulos$^{1,2}$
    {\bf Malvina Nikandrou$^{1}$}
    {\bf Alessandro Suglia$^{1,2}$}\\
    {\bf Oliver Lemon$^{1,2}$}
    {\bf \enspace Arash Eshghi$^{1,2}$}
    \AND \textnormal{$^1$Heriot-Watt University; $^2$Alana AI}\\
    \AND \textnormal {\texttt{\{gmp2000, mn2002, a.suglia, o.lemon, a.eshghi\}}\texttt{@hw.ac.uk}}
}

\begin{document}
\maketitle
\begin{abstract}
    This study explores replacing Transformers in Visual Language Models (VLMs) with Mamba, a recent structured state space model (SSM) that demonstrates promising performance in sequence modeling.
    We test models up to 3B parameters under controlled conditions, showing that Mamba-based VLMs outperforms Transformers-based VLMs in captioning, question answering, and reading comprehension.
    However, we find that Transformers achieve greater performance in visual grounding and the performance gap widens with scale.
    We explore two hypotheses to explain this phenomenon: 1) the effect of task-agnostic visual encoding on the updates of the hidden states, and 2) the difficulty in performing visual grounding from the perspective of in-context multimodal retrieval.    
    Our results indicate that a task-aware encoding yields minimal performance gains on grounding, however, Transformers significantly outperform Mamba at in-context multimodal retrieval.
    Overall, Mamba shows promising performance on tasks where the correct output relies on a summary of the image but struggles when retrieval of explicit information from the context is required\footnote{Code available \href{https://github.com/gpantaz/vl_mamba}{here}.}.
\end{abstract}

% \begingroup
% \def\thefootnote{*}\footnotetext{Equal Contribution, $^1$Code available \href{https://github.com/gpantaz/vl_mamba}{here}}
% \endgroup

\section{Introduction}
% Visually-conditioned language models (VLMs) that provide textual responses to image-text prompts have recently received a growing interest for a variety of applications.
Modern Visual Language Models (VLMs) \citep{bai2023qwenlm, li2024llava, alayrac2022flamingo} typically treat patch representations from vision encoders \citep{radford2021learning, fang2023eva, zhai2023sigmoid} as tokens that are mapped to the embedding space of a Transformer-based Large Language Model (LLM). 
This patch-as-token approach has fostered the development of VLMs that have achieved unprecedented performance on established Vision \& Language (VL) on many coarse-grained tasks, for example, image captioning \citep{lin2014microsoft} or visual question answering \citep{goyal2017making, hudson2019gqa}.
However, fine-grained tasks such as localizing regions within an image \citep{peng2023kosmos, kazemzadeh2014referitgame}, or reading text \citep{sidorov2020textcaps, mathew2021docvqa} from the image are significantly more challenging for these models. 
These tasks require the model to grasp nuances within the image beyond summarizing the visual context in a few words as in conventional image captioning.

A straightforward countermeasure is to scale up the resolution of images, allowing the VLM to ``see greater details''. \citep{liu2023improved, karamcheti2024prismatic, mckinzie2024mm1}.
On the other hand, increasing the context length requires substantial overhead as Transformer-based VLMs have quadratic complexity with respect to the input.
Structured state space models (SSMs) \citep{gu2022efficiently, poli2023hyena} have recently emerged, providing competitive performance against Transformers. Mamba \citep{gu2023mamba} is a recent SSM that promises computational efficiency as well as performance that surpasses Transformer-based language models of similar size.

% In this work, to establish whether the Mamba backbone offers a competitive alternative to Transformers for VLMs, we investigate the performance of a Mamba-based VLM across established multimodal tasks including both fine-grained and more high-level multimodal tasks.

In this paper, we investigate whether a Mamba LLM is a competitive alternative to a Transformer across established multimodal tasks including both fine-grained and coarse-grained multimodal tasks.
The choice of the LLM plays a crucial role for modern VLMs, as recent work \citep{laurenccon2024matters} has shown that for a fixed number of total parameters, the quality of the language backbone has a higher impact than that of the vision backbone. 
% for the same number of parameters, the quality of the language backbone has a higher impact on the final model than the quality of the vision backbone \cite{laurenccon2024matters}.
More specifically, we train three Mamba-VL variants and compare them against Pythia-VL, a series of equally sized models that follow the established paradigm to train VLMs with a state-of-the-art Transformer-based LLM backbone \citep{biderman2023pythia}.
Notably, the performance of Pythia-VL is comparable with that of existing VLMs, thus establishing it as a robust baseline model.
We emphasize that both models are trained on the exact same data presented in the same order, and with identical training hyperparameters, allowing us to provide precise indications of the strengths and weaknesses of the two approaches.

We find that Mamba-VL outperforms Pythia-VL in captioning, question answering, and reading comprehension, but Pythia-VL models consistently achieve greater performance in grounding tasks, and this gap widens in larger models. 
To identify the issue of the difference in performance, we explore the impact of task-agnostic visual encoding, where the model produces embeddings for image representations without information about the task. %, and framing grounding as a in-context multimodal retrieval.
While task-aware image encoding provides a modest improvement in Mamba-VL's grounding capabilities, it remains inferior to the performance of the Transformer-based VLMs.
We investigate this further by casting visual grounding as an in-context multimodal retrieval task, where the model has to retrieve the correct token from the sequence associated with the query. 
Our results show that Transformers are notably more sample efficient, indicating an inherent limitation of Mamba in retrieval-oriented tasks, despite the promising results in sequence modeling.
%both in terms of success rate as well as sample efficiency.
All in all, these experiments showcase that Mamba can be quite effective when the downstream task requires a summary of the image but struggles in tasks where it has to retrieve fine-grained details from the image.

\section{Related Work}
\subsection{VLMs}
Early works showcase the capabilities of LLMs combined with pretrained vision encoders, in VL tasks \citep{tsimpoukelli2021multimodal}.
Consequently, current VLMs \citep{bai2023qwen, dai2024instructblip, alayrac2022flamingo, laurenccon2024matters, liu2024visual, chen2023pali} are based on the same foundational formula: a visual expert \citep{zhai2023sigmoid, fang2023eva}, a language backbone \citep{touvron2023llama, jiang2023mistral, bai2023qwenlm, team2024gemma}, and a connector between the two modules. 
The vast majority of these models are based on highly capable Transformer-based LLMs.
In this work, while we do not modify this formula, we investigate the effect of replacing the Transformer LLM with Mamba.

\subsection{Structured State Space Models}

Structured state space sequence models (S4) are a family of models of sequence models using principles from RNNs, CNNs, and classical state space models that attempt to combat the limitations of Transformers in modeling long sequences \citep{fu2023hungry, poli2023hyena, gu2022efficiently, smith2023simplified}.
These models showcase convincing results in modeling long-range dependencies across several synthetic tasks \citep{tay2021long}.
Previous research shows, in a controlled study of moderately sized models,  that Transformers outperform S4 models in terms of language modeling \citep{arora2024on}.
However,  Mamba \citep{gu2023mamba} builds upon previous S4 models by introducing a selective scan operation (\cref{sec:prelim}) showing competitive performance against Transformers.

\begin{figure*}[tb]
    \centering
    \includegraphics[width=\linewidth]{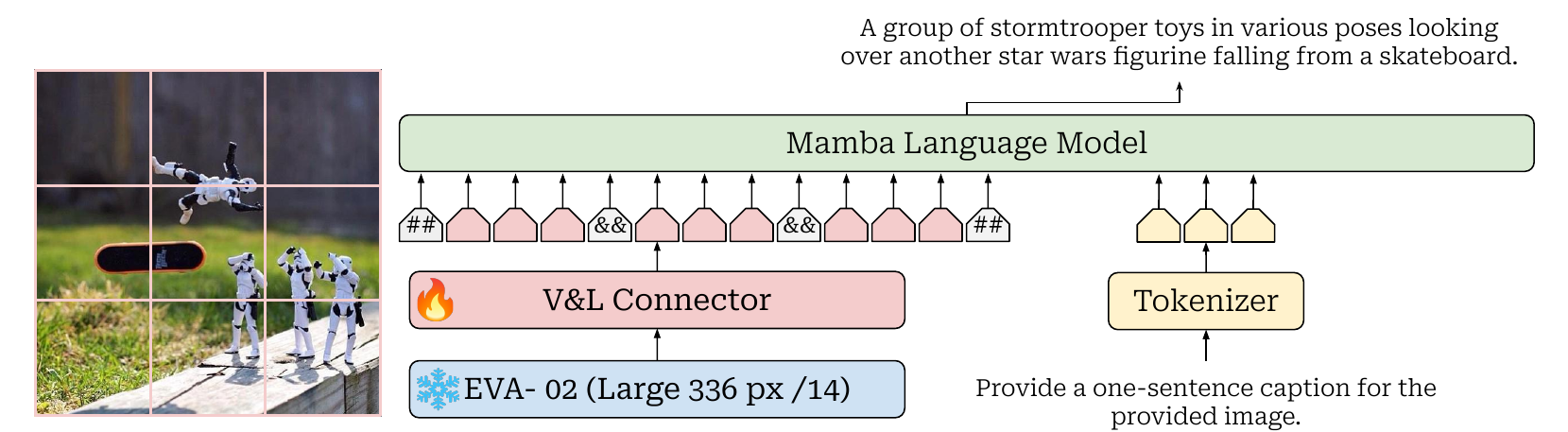}
    \caption{Overview of Mamba-VL. We embed images using EVA-02 and use an MLP as V\&L connector to align the image with text embeddings before the Mamba backbone. Because Mamba does not encode positional information, we introduce custom tokens that delineate the beginning and the end position of the image in the sequence. We also use custom tokens that act as row separators within the image. The vision encoder is kept frozen during training.}
    \label{fig:mamba-model}
\end{figure*}

\paragraph{Mamba applications} 
% Inspired by its results in sequence modeling, concurrent work applies Mamba to other domains. 
% In computer vision, recent work builds upon Mamba with inductive biases such as bidirectionality, positional information \citep{zhu2024vision}, or modifications in the selective scan operation \citep{huang2024localmamba, ruan2024vm, liu2024swin} that better match the domain of image encoding.
Inspired by its results in sequence modeling, recent work applies Mamba to computer vision tasks, by introducing inductive biases that better match the domain of image encoding \citep{zhu2024vision, huang2024localmamba, ruan2024vm, liu2024swin}.
Within NLP, Jamba \citep{lieber2024jamba} is a hybrid architecture with interleaved Transformer and Mamba blocks, while MambaByte \citep{wang2024mambabyte}, is a language model operating on bytes instead of subwords.
To the best of our knowledge, there is not yet a comprehensive study showcasing the effectiveness of Mamba in multimodal settings. 
Concurrent work has applied Mamba in multimodal tasks \citep{zhao2024cobra, qiao2024vl}.
% However, these studies offer limited insights, as there is no control over the training regime, data distribution and model capacity.
However, these studies offer limited insights, because 1) they do not facilitate a fair comparison under controlled conditions, and 2) they do not investigate multimodal tasks that require both high-level and fine-grained information, such as visual grounding. 
% Thus, they offer limited insights into designing robust and versatile VLMs.
% In our study, we create a reproducible experimental setup for comparing these two architectural choices in creating VLMs.

\paragraph{Transformers vs SSMs}

The development of SSMs and similar RNNs \citep{katharopoulos2020transformers, fu2023hungry, peng2023rwkv, poli2023hyena} with competitive performance, has motivated comparisons with Transformers.
Recent studies \citep{park2024can, grazzi2024mamba} show that SSMs can match the in-context learning performance of Transformers on certain tasks, but \citet{akyurek2024context} demonstrate that Transformers retain an advantage for in-context language learning.
Moreover, \citet{merrill2024illusion} provide theoretical and empirical evidence contrary to previous claims \citep{gu2021combining}, showing that SSMs and Transformers have limited expressivity making them unsuitable for state-tracking problems. 
% Their evaluation across state-tracking problems underscores that models like Mamba exhibit similar limitations to Transformers in sequential tasks, but they appear to be better at state-tracking on non-commutative tasks. 
In terms of the in-context retrieval (e.g., copying) capabilities of selective SSMs, \citet{gu2023mamba} show that Mamba is capable of performing associative recall, as formulated by the Induction Heads \cite{olsson2022context} task.
However, follow-up work \cite{jelassi2024repeat, wen2024rnns} provides evidence that SSMs fall behind Transformers when the copying task requires precise retrieval from the context.
% \citet{jelassi2024repeat} showcase that a two-layer Transformer can copy strings of exponential length, while SSMs have limited capabilities due to their fixed-size latent state. 
% These results are further corroborated by their empirical findings, which show that Transformers seem to be better at copying or retrieving part of the input context.  
% compare with induction heads and mamba paper. \cite{gu2023mamba} show that Mamba is capable to perform the associative recall (Induction Heads \cite{Olsson}) task and extrapolate to longer sequences.
% The struggle of SSMs with retrieval tasks may pose significant implications, particularly when the model is required to learn from the input context, such as retrieval-augmented-generation or in-context-learning. 
% Nonetheless, SSMs hold several advantages over Transformers, including constant memory and computational complexity regardless of input length, making them well-suited for handling lengthy inputs. 
We leverage these insights from previous work to draw parallels with VL tasks.
In particular, we formulate a synthetic task for multimodal in-context retrieval to explain the limitation of Mamba in visual grounding.

\section{VLM Approach}

\subsection{Preliminaries: The Mamba model}\label{sec:prelim}
S4 models \citep{gu2022efficiently} take inspiration from Linear Time-Invariant (LTI) models that map a sequence $x(t) \in \mathbb{R} \rightarrowtail y(t) \in \mathbb{R}$ through a hidden state $h(t) \in \mathbb{R}^N$. The output of an LTI model is computed in a two-stage format:

\vspace{-1em}
\begin{gather*}\label{eq:continuous}
h'(t) = Ah(t) + Bx(t),\tag{1a}\\
y(t) = Ch(t)\tag{1b}
\end{gather*}

S4 models first transform the continuous parameters ($\mathbf{A}$, $\mathbf{B}$) with a discretization step with $\mathbf{\Delta}$ parameters, into discrete parameters ($\bar{\mathbf{A}}, \bar{\mathbf{B}}$).
Given the discrete parameters $\mathbf{\bar{A}}, \mathbf{\bar{B}}$ the discrete update is defined in recurrent form \cref{eq:discrete-recurrent}, or via the convolution form \cref{eq:discrete-convolution}:

\vspace{-1em}
\begin{gather*}\label{eq:discrete-recurrent}
h_t = \mathbf{\bar{A}}h_{t-1} + \mathbf{\bar{B}}x_t\tag{2a}\\
y_t = \mathbf{C}h_{t}\tag{2b}
\end{gather*}

\vspace{-1em}
\begin{gather*}\label{eq:discrete-convolution}
\mathbf{\bar{K}} = (\mathbf{C}\mathbf{\bar{B}}, \mathbf{C}\mathbf{\bar{A}}\mathbf{\bar{B}}, \dots, \mathbf{C}\mathbf{\bar{A}}^k\mathbf{\bar{B}})\tag{3a}\\
y = x * \bar{\mathbf{K}}\tag{3b}
\end{gather*}

However, for language modeling S4 models underperform attention-based models \citep{arora2023zoology}.
\citet{gu2023mamba} empirically show that the time-independent parameters of an S4 model are not sufficient to select the correct information from their context as it is not straightforward how to reset the hidden state at each timestep.
For this purpose, certain parameters of the Mamba model ($\Delta, \mathbf{B}, \mathbf{{C}}$) are allowed to be functions of the input. 
With this change, hidden states can be updated in a selective fashion over the input -- though due to violation of the convolution view (\cref{eq:discrete-convolution}), this requires a hardware-aware implementation to compute the hidden states efficiently.
For additional implementation details of Mamba please see the original paper \citep{gu2023mamba}.

\begin{figure*}[tb]
    \centering
    \includegraphics[width=\textwidth]{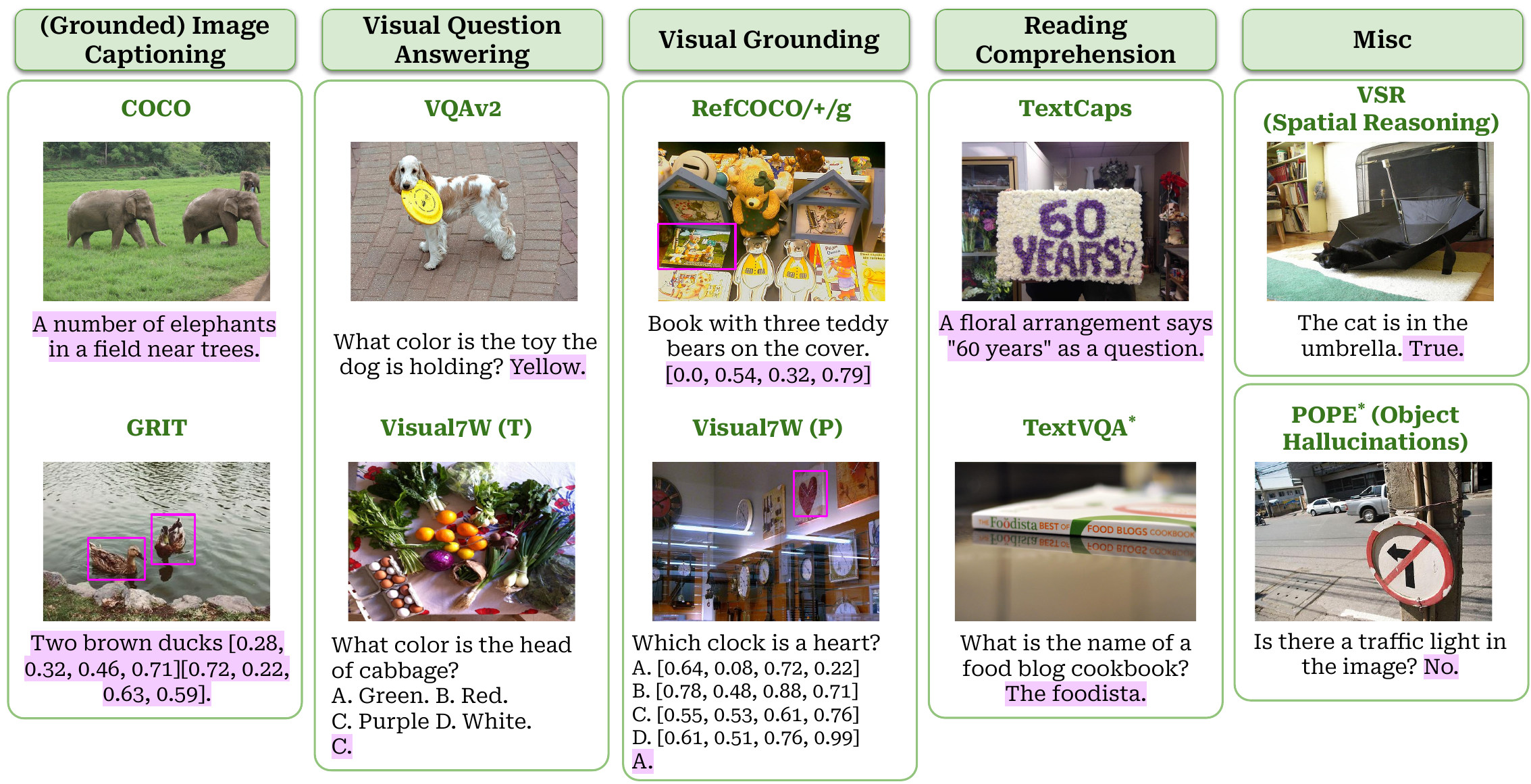}
    \caption{Overview of task categorization and format. We leverage a collection of datasets for coarse-grained (e.g., image captioning, visual question answering) and fine-grained (e.g., visual grounding, reading comprehension) multimodal tasks. \colorbox{lightmagenta}{Text in purple} indicates the outputs of a model for each task. $^*$ denotes held-out datasets.}
    \label{fig:all_tasks}
\end{figure*}

\subsection{Model Architecture} 
\cref{fig:mamba-model} shows an overview of our model. 
We built our approach using the standard paradigm for VLMs that combine unimodal experts \citep{liu2024visual, alayrac2022flamingo, dai2024instructblip}. 
More specifically, our model consists of three individual components, a vision encoder, the Vision \& Language connector, and the language backbone.

\paragraph{Vision Encoder} We use EVA-02-L336px/14 \citep{fang2023eva} to obtain high-quality visual representations. 
While previous work usually adopts CLIP models \citep{bai2023qwen, liu2024visual}, the EVA series outperforms the existing open CLIP models. 
We also provide results in \cref{appendix-experiments}, showcasing a comparison between the two vision encoders using preliminary checkpoints. 
Furthermore, based on previous work \citep{karamcheti2024prismatic}, we opted for higher resolution images, as it has been shown that it leads to performance gains.

\paragraph{Vision \& Language Connector} We follow LLaVA-1.5 \citep{liu2023improved} and use a two-layer MLP that projects the visual tokens to the dimensionality expected by the LLM, leaving more sophisticated architectural choices \citep{dai2024instructblip, bai2023qwen, you2023ferret} for future work.

\paragraph{Language Backbone} 
We use Mamba or Pythia \cite{biderman2023pythia} as the language backbone that accepts the visual features from the connector module, and the tokenized text containing the task instruction and any sample text.
We select Pythia as the baseline Transformer-based language model because it enables direct comparison as it 1) follows the state-of-the-art Transformer recipe \citep{su2024roformer, dao2023flashattention}, 2) is trained on the same dataset as Mamba \citep{gao2020pile}, 3) provides model variants with a similar number of parameters.

A key difference between the two models is that Mamba does not allocate parameters to model positional information.
This inductive bias has been identified by concurrent work \citep{liu2024vmamba, zhu2024vision}, applying Mamba to computer vision tasks, since positional embeddings capture the structure of the image. 
Inspired by Fuyu \citep{Fuyu}, we overcome this issue by introducing a separator token (``\#\#'') that signals the beginning and the end of the image sequence, as well as an image-newline character (``\&\&'') that depicts the end of a row of patches.

\begin{table*}[tb]
    \centering
    \scriptsize
    % \small
    \renewcommand{\arraystretch}{1.2}
    % \begin{adjustbox}{center}
        \begin{tabular}{@{}ll |c cl| c c cl| c c @{}} 
            \toprule
            & & \multicolumn{3}{c|}{\textbf{Image Captioning}} & \multicolumn{4}{c|}{\textbf{General VQA}} & \multicolumn{2}{c}{\textbf{Misc}}\\
            Model & LLM  & COCO & NoCaps$^*$ & \multicolumn{1}{c|}{\underline{Sum}} & VQAv2 & GQA &  V7W & \multicolumn{1}{c|}{\underline{Sum}} & VSR & POPE$^*$ \\
            & Param & test & val & & val & test-dev & test-T & &  test & test \\
            \midrule
            Pythia-VL & 1B & 132.89 & 97.61 & 230.50 & 72.26 & 53.79 & 81.96 & 208.81 & 72.43 & 86.77\\
            Mamba-VL & 790M & 133.81 & 99.00 & 232.81 \textcolor{mygreen}{\scriptsize{(+2.31)}} & 71.67 & 54.95 & 81.82 & 208.44  \textcolor{myred}{\scriptsize{(-0.37)}}& 75.39 & 86.77\\
            \midrule
            Pythia-VL & 1.4B & 134.06 & 100.72 & 234.78 & 73.57 & 57.05 & 83.06 & 213.68 & 77.72 & 86.40\\
            Mamba-VL & 1.4B & 134.76 & 100.56 & 235.32 \textcolor{mygreen}{\scriptsize{(+0.54)}}& 74.46 & 58.44 & 83.78 & 216.67 \textcolor{mygreen}{\scriptsize{(+2.99)}}& 80.18 & 85.32\\
            \midrule
            Pythia-VL & 2.8B & 134.97 & 101.27 & 236.24 & 75.08 & 59.76 & 84.34 & 219.18 & 80.86 & 86.87\\
            Mamba-VL & 2.8B & \textbf{135.53} & \textbf{102.00} & \textbf{237.53} \textcolor{mygreen}{\scriptsize{(+1.29)}} & \textbf{76.08} & \textbf{60.41} & \textbf{85.31} & \textbf{221.80}  \textcolor{mygreen}{\scriptsize{(+2.62)}}& \textbf{81.45} & \textbf{87.33}\\
            \bottomrule
        \end{tabular}
    % \end{adjustbox}
    \caption{Results on image captioning, general VQA, and misc benchmarks. $^*$ denotes zero-shot performance.}
    \label{tab:coarse-grained-results}
\end{table*}

\begin{table*}[tb]
    \centering
    \scriptsize
    % \small
    \renewcommand{\arraystretch}{1.2}
    \addtolength{\tabcolsep}{-0.3em}
    % \begin{adjustbox}{center}
    % \resizebox{\linewidth}{!}{
        \begin{tabular}{@{}ll |ccccccl| cccl@{}} 
            \toprule
            % Model & Captioning & General VQA & Ref Expr \\ 
            & & \multicolumn{7}{c|}{\textbf{Visual Grounding}} & \multicolumn{4}{c}{\textbf{Reading Comprehension}}\\
            % Model & LLM  & \multicolumn{5}{c}{RefCOCO/+/g} & V7W & \multicolumn{1}{c|}{\underline{Sum}} & TextCaps & TextVQA$^*$ & AI2D & \multicolumn{1}{c}{\underline{Sum}}\\
            Model & LLM  & \multicolumn{2}{c}{RefCOCO} & \multicolumn{2}{c}{RefCOCO+} & RefCOCOg & V7W & \multicolumn{1}{c|}{\underline{Sum}} & TextCaps & TextVQA$^\dagger$ & 
            AI2D & \multicolumn{1}{c}{\underline{Sum}}\\
            & Param & test-A & test-B & test-A & test-B & test & test-P & & val & val & test\\
            \midrule
            Pythia-VL & 1B & 76.00 & 62.48 &  45.36 & 47.44 & 67.58 & 83.78 & 382.64 & 92.73 & 35.22 & 77.62 & 205.57\\
            Mamba-VL & 790M & 67.84 & 56.35 & 57.97 & 41.43 & 59.16 & 74.01 & 356.76 \textcolor{myred}{\scriptsize{(-25.88)}} & 94.30 & 40.72 & 79.27 & 214.29 \textcolor{mygreen}{\scriptsize{(+8.72)}}\\
            \midrule
            Pythia-VL & 1.4B & 82.43 & 68.39 & 72.35 & 55.16 & 72.56 & 86.13 & 437.02 & 94.60 & 37.54 & 79.27 & 211.41\\
            Mamba-VL & 1.4B & 76.60 & 63.48 & 68.40 & 52.11 & 68.82 & 80.18 & 409.59 \textcolor{myred}{\scriptsize{(-27.43)}}& 98.68 & 41.30 & 80.86 & 220.84 \textcolor{mygreen}{\scriptsize{(+9.43)}}\\
            \midrule
            Pythia-VL & 2.8B & \textbf{85.39} & \textbf{70.82} & \textbf{75.39} & \textbf{58.62} & \textbf{76.24} & \textbf{86.61} & \textbf{453.07} & 99.74 & 39.14 & 81.57 & 220.45\\
            Mamba-VL & 2.8B & 79.29 & 64.97 & 71.64 & 53.94 & 71.27 & 82.50 & 423.61 \textcolor{myred}{\scriptsize{(-29.45)}} & \textbf{100.47} & \textbf{42.14} & \textbf{83.71} & \textbf{226.32} \textcolor{mygreen}{\scriptsize{(+5.87)}}\\
            \bottomrule
        \end{tabular}
    % }
    % \end{adjustbox}
        \caption{Results on visual grounding, and text-oriented, benchmarks. $\dagger$ denotes a task not in the training mixture.}
    \label{tab:fine-grained-results}
\end{table*}
\section{Datasets}
% Our goal is to facilitate a fully reproducible comparison between VLMs with Transformer and Mamba backbones.
% We exclusively focus on datasets that are fully open-source and have been applied to train previous models to facilitate a fully reproducible comparison of VLMs with Transformer and Mamba backbones.
% For this purpose, we use two datasets, for pretraining and instruction tuning respectively.
% We use a collection of open-source datasets that allow for a fully reproducible comparison.
% \paragraph{Pretraining} We leverage the dataset from \citet{liu2024visual}, a subset of 595K captions from Conceptual Captions 3M \citep{sharma2018conceptual}.
% %, to achieve a good balance between training efficiency and concept coverage.
% \paragraph{Instruction-Tuning} 
% The second dataset is a compilation of existing datasets and is used for instruction tuning.
We use a collection of open-source datasets to allow a fully reproducible comparison.
For pretraining, we leverage the dataset from \citet{liu2024visual}, a subset of 595K captions from Conceptual Captions 3M \citep{sharma2018conceptual}.
For instruction tuning, we use a collection of established coarse and fine-grained vision-language tasks (e.g., captioning, visual question answering, and referring expression).
\cref{fig:all_tasks} shows examples for all tasks in our training and evaluation.
% \cref{tab:instruction-tuning-datasets} shows the datasets used for instruction tuning.
We provide details for our dataset, filtering approach, and task instructions in \cref{sec:appedix-dataset}.
Notably, we pack the examples from the same image and task into one sequence.
% In total, our model is trained on 6.2M packed examples.
% The benefits of this approach are two-fold: 1) we ensure efficiency in training since padding is minimized \citep{krell2021efficient}, and 2) by packing examples we facilitate chat capabilities of our models to some degree.

\section{Experiments}
% \subsection{Baselines}
% We compare Mamba-VL against strong baseline Transformer models by using Pythia \citep{biderman2023pythia} as the LLM backbone of the VLM.
% Pythia is a model trained on the same dataset as Mamba \citep{gao2020pile} with similar hyperparameters and follows the state-of-the-art recipe of a Transformer \citep{su2024roformer, dao2023flashattention}.

\subsection{Experimental Setup} Similar to previous work \citep{liu2024visual, li2024llava} we employ a two-step training regime. 
First, we perform a warmup stage where we train only the connector component on the pretraining dataset.
Next, we unfreeze the language model parameters and train on the instruction-tuning dataset. 
All models are trained using the same data, in the same order, and with identical training hyperparameters (see \cref{appendix:training-details} for further details).
% In this way, we can facilitate a fair comparison between the two LLM backbones across different model sizes.
% Additionally, we also provide comparison with state-of-the-art publicly available VLMs in \cref{sec:exp-sota}.
% During this training stage, we evaluate the model every 10k steps.
% \cref{appendix:training-details} shows information regarding training details from both stages.
Unless stated otherwise, we report the evaluation performance without task-specific fine-tuning.
% The evaluation metrics for each dataset are listed in \cref{tab:benchmark_metrics}.

\subsection{Results}\label{sec:results}
% \begin{enumerate}
%     \item Because the hidden state is updated based on solely the previous state and the current timestep, Mamba has to perform a "task" agnostic visual encoding
%     \item That is not the case in Pythia where the current hidden state attends to every other token in the sequence
% \end{enumerate}

% \begin{gather*}\label{eq:xx}
% \mathbf{\bar{A}}_k = exp(\Delta_k\mathbf{\bar{A}})\tag{4}\\
% \mathbf{\bar{B}}_k = (\mathbf{\bar{A_k}} - 1) (\mathbf{\bar{B}}(t) / A)\tag{4b}
% \end{gather*}

% \begin{itemize}
%     \item Ignore: Delta goes to 0 for tokens that we would like to ignore entirely. If Delta goes to 0 Ak goes to 1 and Bk goes to 0 telling the model that the current token is extremely not important
%     \item Reset: Dleta goes to +00 Ak goes to zero which basically resets the hidden state.
    
% \end{itemize}

\begin{table*}[tb]
    \centering
    % \scriptsize
    \scriptsize
    \renewcommand{\arraystretch}{1.2}
    % \begin{adjustbox}{center}
        \begin{tabular}{@{}l| l|c|cc|cc|cc|c@{}} 
        \toprule
        Model & LLM & NoCaps$^*$ & VQA & GQA & RefCOCOg & V7W (P) & TextVQA     & AI2D & POPE$^*$\\
        & & val & test-dev & test-dev & test & test & val & val & test\\
        \midrule
        \textcolor{gray}{LLaVA-1.5 \shortcite{liu2024visual}} & \textcolor{gray}{Vicuna-7B} & - & \textcolor{gray}{78.5} & \textcolor{gray}{62.0} & - & - & \textcolor{gray}{58.2} & - & \textcolor{gray}{85.8}\\
         \textcolor{gray}{InstructBLIP \shortcite{dai2024instructblip}} & \textcolor{gray}{Vicuna-7B} & \textcolor{gray}{123.1} & - & \textcolor{gray}{49.2} & - & - & \textcolor{gray}{50.1} & - & \textcolor{gray}{83.7}\\
         \textcolor{gray}{Shikra \shortcite{chen2023shikra}} & \textcolor{gray}{Vicuna-7B} & - & \textcolor{gray}{77.4} & - & \textcolor{gray}{82.19} & \textcolor{gray}{85.33} & - & - & \textcolor{gray}{83.9}\\
         \textcolor{gray}{Ferret-v2-7B \shortcite{zhang2024ferret}} & \textcolor{gray}{Vicuna-7B} & - & \textcolor{gray}{81.5} & \textcolor{gray}{64.7} & \textcolor{gray}{89.27} & - & \textcolor{gray}{61.7} & - & \textcolor{gray}{87.8}\\
         \textcolor{gray}{Qwen-VL-Chat \shortcite{bai2023qwen}} & \textcolor{gray}{Qwen-7B} & \textcolor{gray}{120.2} & \textcolor{gray}{78.2} & \textcolor{gray}{57.5} & \textcolor{gray}{86.32} & & \textcolor{gray}{61.5} & \textcolor{gray}{62.3} & -\\
         \textcolor{gray}{IDEFICS2 \shortcite{laurenccon2024matters}} & \textcolor{gray}{Mistral-7B-v0.1} & - & \textcolor{gray}{81.2} & - & - & - & \textcolor{gray}{73.0} & - & - \\
        \midrule
        LLaVA-Phi \shortcite{zhu2024llava} & Phi2-2.7B & - & 71.4 & - & - & - & 48.6 & - &  85.0\\
        TinyLLaVA \shortcite{zhou2024tinyllava} & Phi2-2.7B & - & 79.9 & 62.0 & - & - & 59.1 & - & 86.4\\
        Cobra \shortcite{zhao2024cobra} & Mamba-2.8B & - & 75.9 & 58.5 & - & - & 46.0 & - & 88.0\\
        VL-Mamba \shortcite{liu2024vmamba} & Mamba-2.8B & - & 76.6 & 56.2 & - & - & 48.9 & - & 84.4\\
        \midrule
        Pythia-VL & Pythia-2.8B & 100.72 & 77.0 & 59.8 & 76.24 & 86.61 & 39.1 & 81.6 & 86.9\\
        Mamba-VL & Mamba-2.8B & 100.56 & 78.0 & 60.4 & 71.27 & 82.50 & 42.1 & 87.3 & 87.3\\
        \bottomrule
        \end{tabular}
    \caption{Results against state-of-the-art models. $^*$ denotes zero-shot performance.}
    \label{tab:sota-results}
\end{table*}

\begin{figure}[tb]
    \centering
    \includegraphics[width=\linewidth]{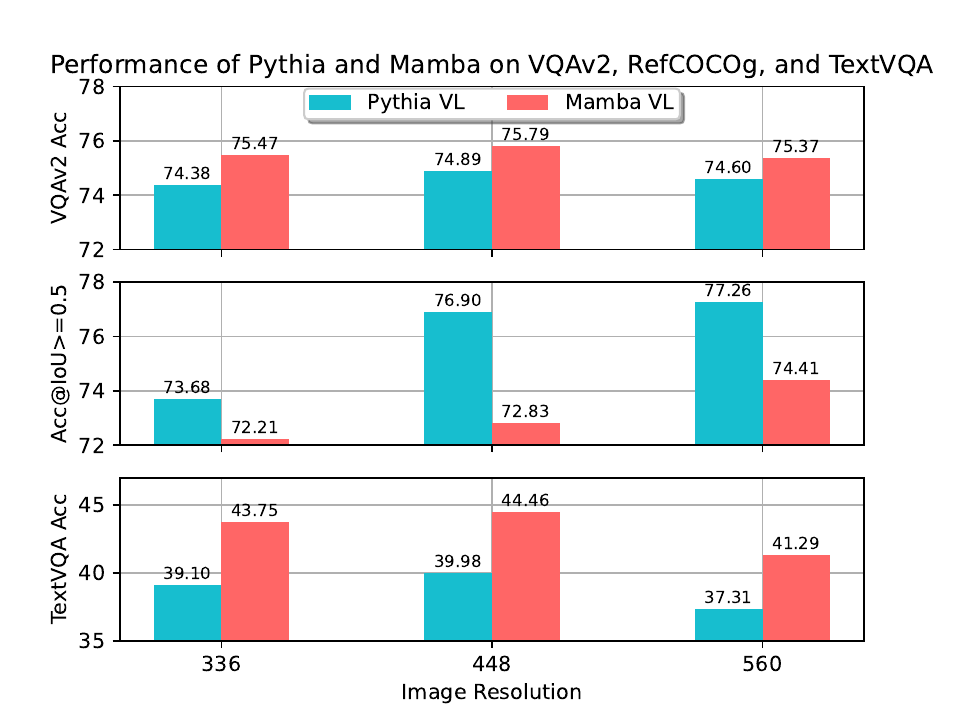}
    \caption{Results of finetuned 1.4B models with increased resolution on VQAv2 (top), RefCOCOg (middle), and TextVQA (bottom). Increasing the resolution to 480$\times$480 pixels results better performance for both models, however, Pythia benefits significantly more than Mamba in the grounding task.}
    \label{fig:finetune_results_hd}
\end{figure}

\paragraph{Pythia vs Mamba}
\cref{tab:coarse-grained-results} and \cref{tab:fine-grained-results} illustrate the comparison between Pythia-VL and Mamba-VL across three model sizes.
We provide results for each benchmark individually, along with a summation score as an indication of overall performance for a task group.
We observe that Mamba variants match or surpass the performance of models with Pythia as an LLM across all three sizes in most tasks.
Specifically, the smallest Mamba-VL achieves competitive performance with Pythia-VL even though it has approximately 200M fewer parameters but also outperforms Pythia-VL on zero-shot image captioning (NoCaps) and on spatial understanding (VSR).
However, the performance gap decreases proportionally to the size of the compared models.
The largest performance difference is observed in the reading comprehension tasks.
We hypothesize that textual information within an image provides a strong signal for Mamba to maintain this information in the hidden state.
Surprisingly, Pythia-VL models consistently outperform Mamba-VL on grounding tasks across all scales, but also this gap is further widened in larger models.

\paragraph{Finetuning with Higher Resolution}
It is widely known that increasing the image resolution yields benefits in Transformer-based VLMs \citep{karamcheti2024prismatic, laurenccon2024matters}.
We explore whether the benefits of higher image resolution translate to Mamba given its strong long sequence modeling capabilities \citep{gu2023mamba}.
\cref{fig:finetune_results_hd} shows the performance of 1.4B models on VQAv2, RefCOCOg, and TextVQA after finetuning on each task with higher-resolution images.
As expected, both models benefit from higher-resolution images, and the differences are more evident in RefCOCOg, possibly due to the granularity of the task.
Comparing Pythia-VL and Mamba-VL, both models exhibit a similarly small improvement in VQAv2 and TextVQA, but Pythia-VL benefits substantially more than Mamba-VL in RefCOCOg.
This provides further evidence regarding the limitations of Mamba on grounding tasks, on which we further elaborate in \cref{sec:vg}.

\paragraph{Comparison with SOTA models} For completeness, we provide a comparison against state-of-the-art 3B and 7B parameter models (\cref{tab:sota-results}).
We observe that our largest models are competitive even against the largest VLM models. 
Our model performs on par with other Mamba-based VLMs (Cobra and VL-Mamba) with a small advantage in general VQA benchmarks (VQA, GQA).
Importantly, we want to note that our base LLMs have not been instruction-tuned, which could have a major impact, particularly in multimodal language modeling tasks \citep{laurenccon2024matters}. 
Furthermore, it is hard to draw definite conclusions between different models as they have been trained using different datasets and training regimes.  %the data used in training are not the consistent.

\subsection{Why is Grounding Difficult for Mamba?}\label{sec:vg}
We observed that Mamba models are quite effective in multimodal language modeling tasks (e.g., captioning, visual question answering). However, they underperform compared to Transformers of equal capacity in visual grounding tasks.
\emph{What is the underlying reason for this weakness?} 
We explore two possible explanations using the 1.4B parameter models by 1) examining the effect of task-agnostic visual encoding, and 2) framing visual grounding as an in-context multimodal retrieval task. 

% ---- ALE STOPPED HERE----

% \begin{figure}[tb]
%     \centering
%     \includegraphics[width=\linewidth]{figures/task_agnostic.pdf}
%     \caption{Performance of Mamba 1.4B on Grounding benchmarks when the instruction is placed before and after the image representations in the sequence. The yaxis is scaled by the offset of the performance of the resumed checkpoint.}
%     \label{fig:task_agnostic}
% \end{figure}

% TODO: rerun from pretraining checkpoint 2 epochs
\subsubsection{Task-agnostic Visual Encoding} 
\begin{figure}[tb]
    \centering
    \includegraphics[width=\linewidth]{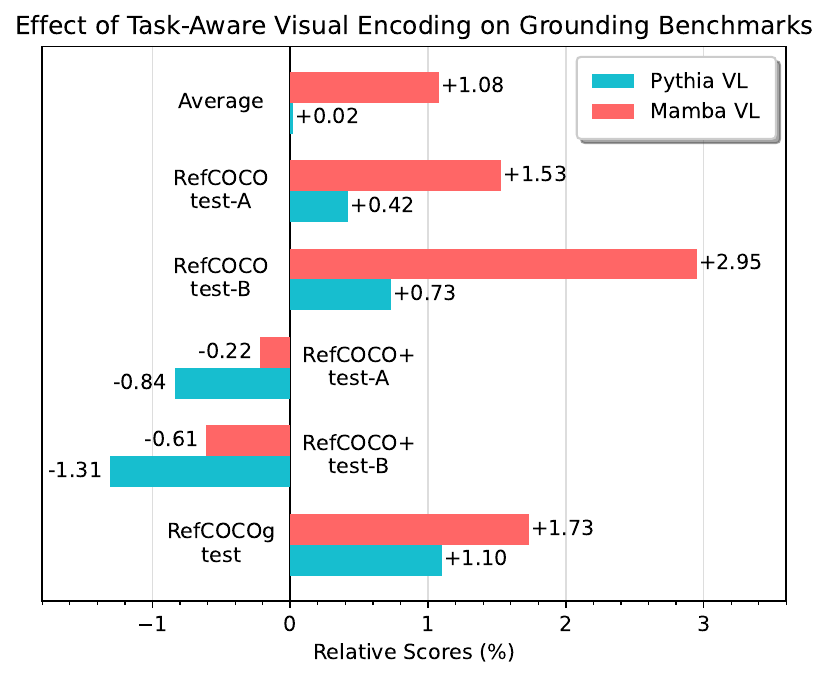}
    \caption{Relative performance difference on visual grounding benchmarks between task-aware and task-agnostic visual encoding. On average, task-aware encoding yields a marginal performance boost on Mamba-VL while it has almost no effect on Pythia-VL.}
    \label{fig:task-aware-grounding-comparison}
\end{figure}

Both Transformer causal models and SSMs operate unidirectionally, i.e. the representation at a given timestep is a function of only the previous and current tokens. 
However, SSMs enforce a stricter update rule, where the hidden state can only be updated with information from the previous hidden state and the current input (\cref{eq:discrete-recurrent}).
Consequently, when the image precedes the instruction, patch representations are encoded in a \emph{task-agnostic} manner.
Intuitively, this might lead the model to store ``generic'' information in its hidden state, which is useful for multimodal language modeling tasks but ineffective in explicit visual grounding, where the model has to remember the spatial positions of any entity in the image.
%Importantly, this is also a property of autoregressive Transformers. 
On the other hand, in Transformer models, the hidden state of each timestep has direct access to all previous timesteps and, therefore, can retrieve relevant information in later hidden states.
%even if the model fails to capture necessary information for a task and the current timestep this information can be incorporated in the next hidden state.

We investigate the impact of \emph{task-aware} visual encoding by placing the task instruction before the image during the instruction-tuning stage. 
In principle, this simple modification favors Mamba as the model may choose to store or ignore inputs that are not relevant to the task. 
\cref{fig:task-aware-grounding-comparison} shows the gain of both models using task-aware encoding on visual grounding benchmarks.
%shows a comparison on grounding tasks after repeating the training stages (see \cref{appendix:task-aware} for full results) where the task instruction precedes the image.
We observe that on average the task-aware encoding leads to a small relative improvement for Mamba-VL, but that even in this setup, Pythia-VL achieves higher performance (see \cref{tab:task-agnostic-full} for full results).
Furthermore, the results vary across different grounding benchmarks, but also across other tasks suggesting that the task-aware encoding is not always beneficial.

Perhaps the performance of Mamba-based VLMs on grounding, as well as on other tasks, can be further improved by incorporating the task instruction, and the query (e.g., question, referring expression) before the image.
This is in line with recent work \citep{jelassi2024repeat} showing that, when the query is available at the beginning of the input, SSMs can perform on par with Transformers on toy associative recall tasks. 
However, this is at odds with the common practice of data packing 
% due to the data packing 
in current VLM training \citep{bai2023qwenlm, li2024llava}. 
We anticipate that naively separating the queries and outputs with image tokens can negatively affect the capabilities of a model.
% We anticipate that this can negatively affect the capabilities of a model, as all instructions need to precede the image tokens, while the outputs have to succeed them.  

% It is worth noting that this problem was somehow mitigated by InstructBLIP~\cite{dai2024instructblip}, a Transformer-based VLM that adopts a connector module able to produce task-aware visual representations of the visual content. This architectural choice might justify the need for more suitable and versatile multimodal fusion architectures. 
% \begin{itemize}
%     \item \citep{dai2024instructblip} showed that the task-aware visual encoding is beneficial in InstructBLIP, a Transformer-based VLM. However, they have opted for a specific architectural choice, where the connector module between the LLM and the vision encoder is producing a task-aware visual prompt that then prepended as a prefix to the instruction for the LLM. This means that in practice the LLM sees first the visual prompt then the instruction/task.
% \end{itemize}

% \begin{figure*}[!htb]
%     \centering
%     \begin{minipage}{.5\textwidth}
%         \centering
%         \includegraphics[width=\linewidth]{figures/Copy of Untitled drawing (10).pdf}
%         % \caption{$dt=0.1$}
%         \label{fig:prob1_6_2}
%     \end{minipage}%
%     \begin{minipage}{0.5\textwidth}
%         \centering
%         \includegraphics[width=\linewidth]{figures/Copy of Copy of Untitled drawing (3).pdf}
%         % \caption{$dt =$}
%         \label{fig:prob1_6_1}
%     \end{minipage}
%     \caption{This is}
%     \label{fig:relation-to-copying}
% \end{figure*}

\subsubsection{Grounding as Multimodal Retrieval}
\begin{figure}[tb]
    \centering
    \includegraphics[width=\linewidth]{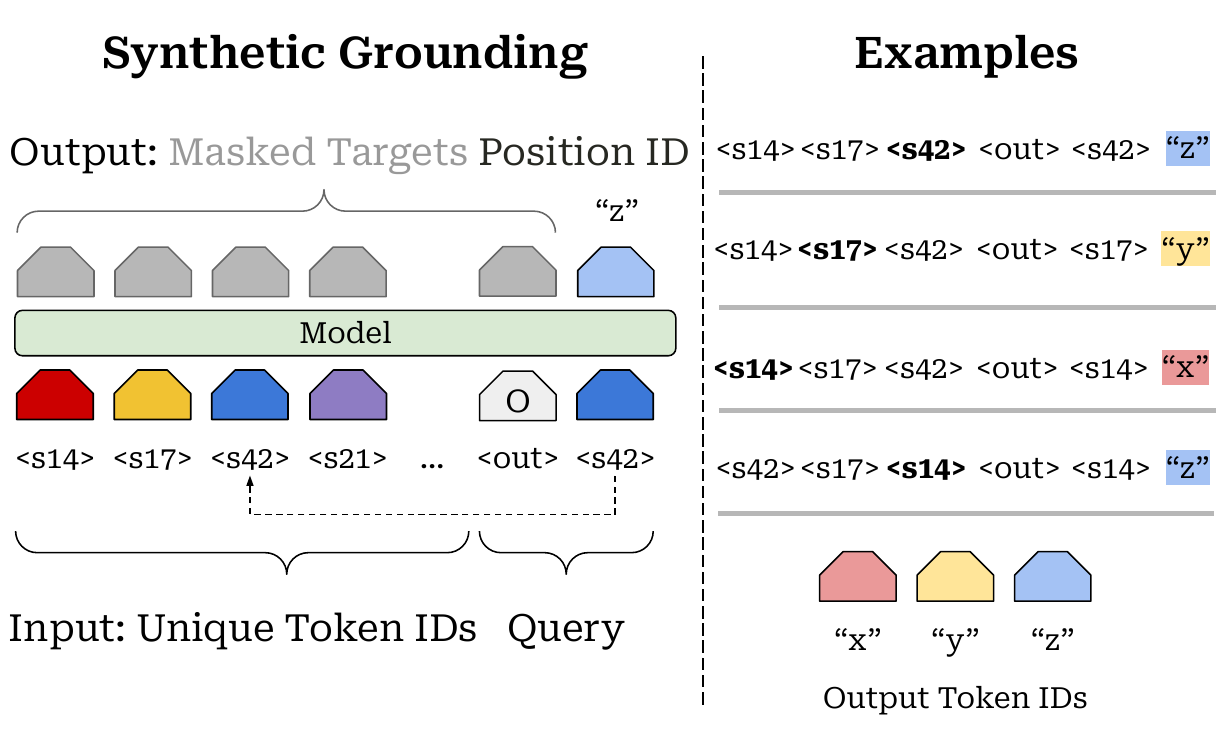}
    \caption{Overview of the synthetic visual grounding task. The model accepts as input a sequence of unique special tokens, followed by an output token and a special token id that appears in the context as a query. The model needs to predict the token id corresponding to the position of the queried token.}
    \label{fig:synthetic_grounding}
\end{figure}

\begin{figure*}[tb]
\minipage{0.33\textwidth}
  \includegraphics[width=\linewidth]{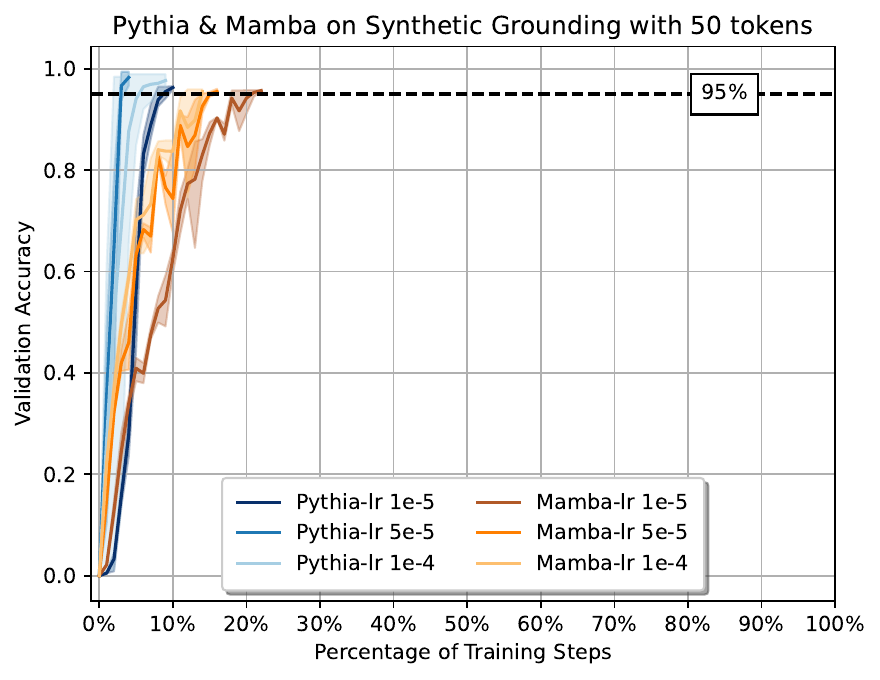}
  \subcaption{Sequence length = 50.}
\endminipage\hfill
\minipage{0.33\textwidth}
  \includegraphics[width=\linewidth]{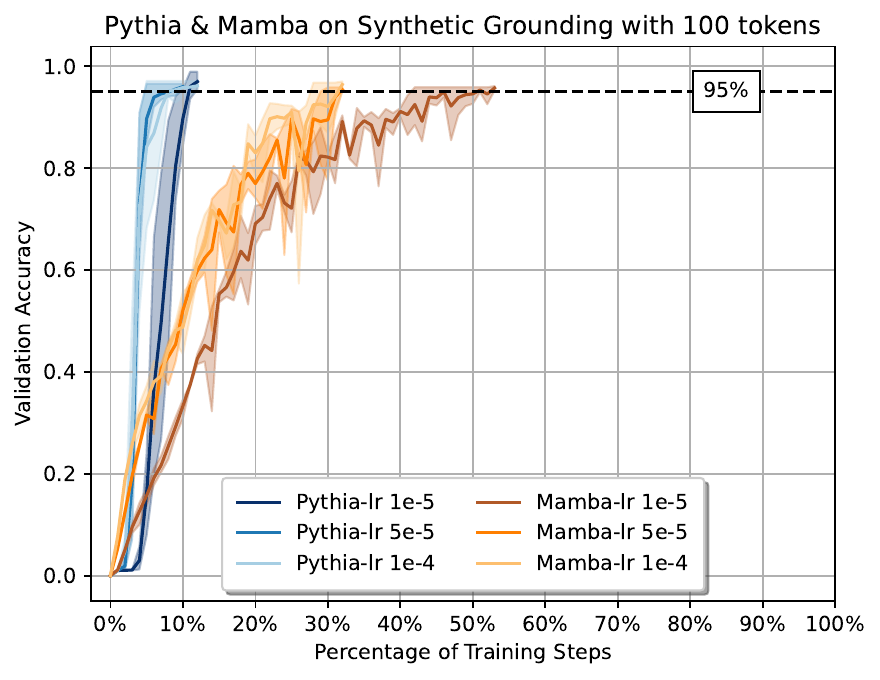}
  \subcaption{Sequence length = 100.}
\endminipage\hfill
\minipage{0.33\textwidth}%
  \includegraphics[width=\linewidth]{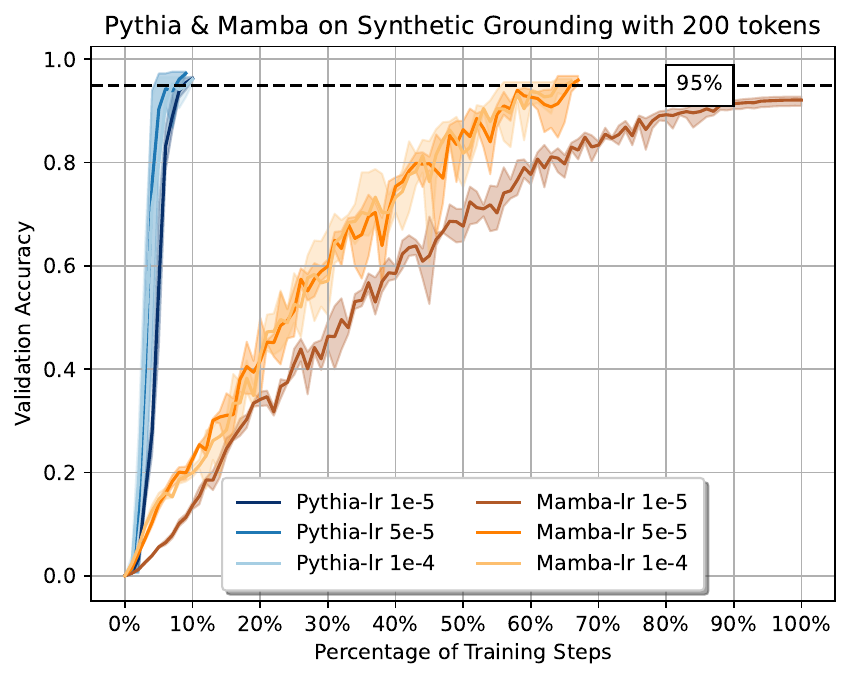}
  \subcaption{Sequence length = 200.}
\endminipage
    \caption{Performance curves for Pythia-1.4B and Mamba-1.4B variants on the synthetic grounding task with varying sequence length. Pythia learns the task significantly faster compared to Mamba.}
    \label{fig:synthetic_grounding_results}
\end{figure*}

% \begin{itemize}
%     \item Draw parallels with copying task
%     \item In copying you need to remember the entire sequence and then after you see the \texttt{<copy>} token you need to output the sequence in the same order
%     \item In Visual Grounding you need to remember the positions of all entities in the image, and once you see the instruction/prompt, you need to output the xyxy coordinates (aka multimodal retrieval)
%     \item Transformers seem to be better at copying than SSMs \citep{jelassi2024repeat} at least on artificial tasks
%     \item Perhaps we can draw the same parallel here, Transformers will be better in visual grounding
%     \item Packing may also have a detrimental effect in this case, because the target tokens move further and further away from the image representation
% \end{itemize}

\begin{figure}[ht!]
\minipage{0.45\textwidth}
  \centering
  \includegraphics[width=0.8\linewidth]{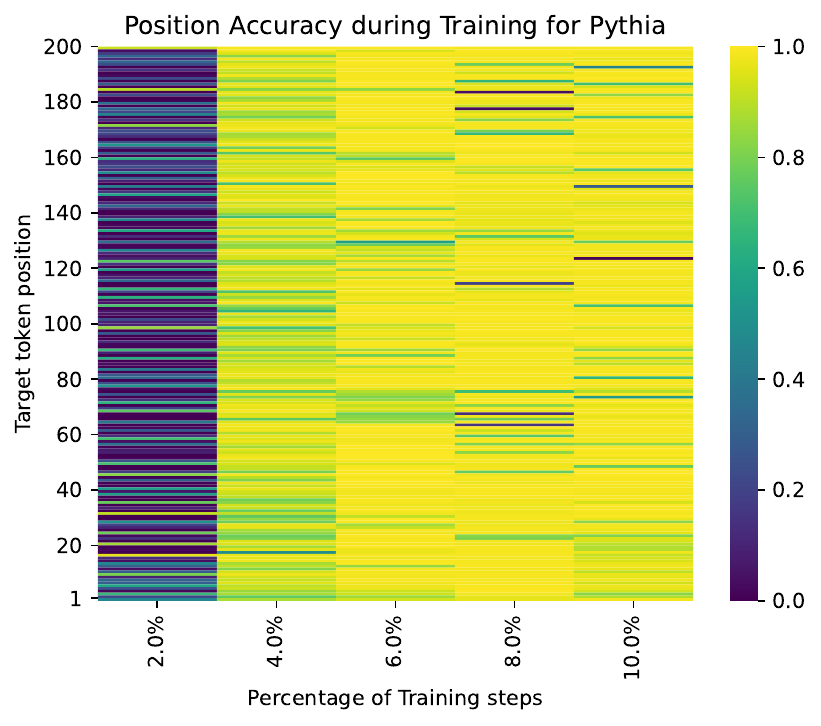}
  % \subcaption{Pythia.}
\endminipage\hfill
\minipage{0.45\textwidth}
  \centering
  \includegraphics[width=0.8\linewidth]{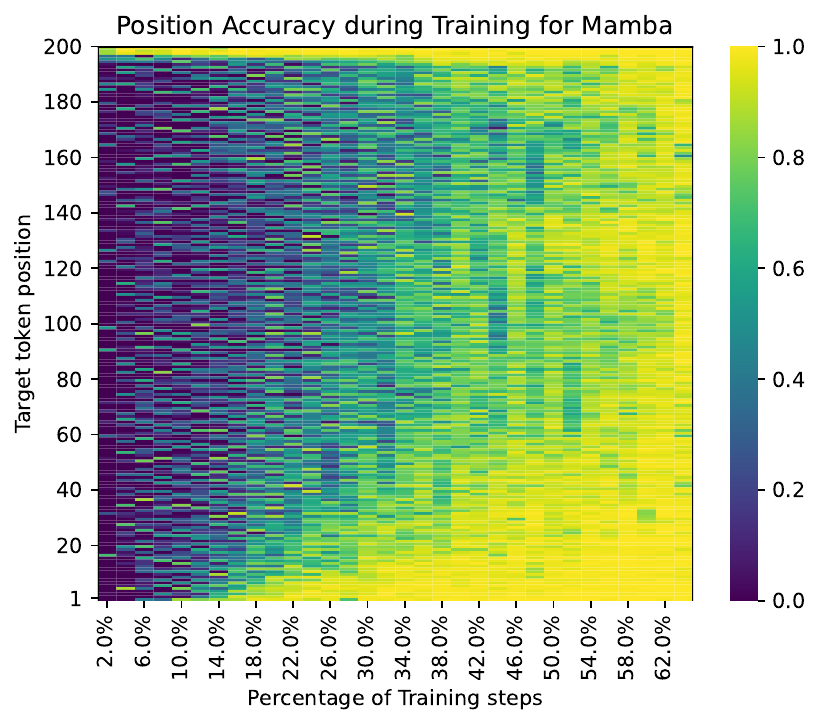}
  % \subcaption{Mamba.}
\endminipage
    \caption{Accuracy per position on the held-out set during training on sequences of 200 tokens.}
    \label{fig:synthetic_grounding_evolution}
\end{figure}

We can view visual grounding as an in-context multimodal retrieval task.
In a standard in-context retrieval task, the model is provided with a context (a text paragraph) and a query (a relevant question), and it needs to extract and copy the part of the input corresponding to the question.
Similarly, in a visual grounding task, the model is provided with a series of patch tokens as context and a prompt and needs to reference the area that matches the prompt.
The core difference is that the space of the token embeddings is different. 
In the standard retrieval task, the inputs and outputs of the model are both in text form, whereas in visual grounding the VLM performs a two-hop step by matching the text prompt to the visual modality and then providing a textual response.

%For this purpose, we introduce a synthetic visual grounding task that is motivated by previous work \citep{jelassi2024repeat, merrill2024illusion}. 
% Motivated by previous work \citep{jelassi2024repeat, merrill2024illusion}, we introduce a synthetic visual grounding task to study this problem (see \cref{fig:synthetic_grounding} for an example). 
For this purpose, motivated by concurrent work \citep{jelassi2024repeat, merrill2024illusion}, we introduce a synthetic task (\cref{fig:synthetic_grounding}) that frames visual grounding as a retrieval objective and facilitates an interpretable model comparison.
%\cref{fig:synthetic_grounding} illustrates our task design with exemplary inputs and outputs. 
We provide a pretrained model with a context of unique special tokens (\texttt{<s14><s17>}\texttt{<s42>}\dots), followed by a query (\texttt{<out><s42>}).
To incorporate the two-hop step between two modalities, we ask the model to return the token id from the vocabulary that matches the position of the special token in the sequence (\texttt{``z''}).
This setup resembles how the language model of a VLM adapts to two modalities. 
We resize the embedding layer of the pretrained models to accommodate the new special tokens (the patch tokens in the VLMs\footnote{In practice, VLMs do not resize the embedding layer of the LLM, but accept the embeddings from the visual encoder.}), and task the model to learn a mapping between the textual and the new embedding space.
Finally, visual grounding is an instantiation of this synthetic task, where the input sequence is composed of the patch representations, the query token is the prompt, and the outputs are the pixels that match the prompt in the image.

We experiment with varying the sequence length (50/100/200, see \cref{appendix:synth_vg} for details). 
For each sequence, we use three different learning rates and train each model with three initializations of the embeddings of the special tokens (9 runs in total per sequence length). 
We track the performance on a held-out set and terminate training whenever the model achieves $\ge95$\% accuracy.

\cref{fig:synthetic_grounding_results} shows the results of both models.
We observe that Pythia learns to solve the task consistently using approximately 10\% of the training data.
On the other hand, Mamba is less sample-efficient requiring nearly double the amount of training when increasing the sequence length, and even fails to reach the accuracy threshold for some runs with longer sequences.
These results show that in tasks requiring access to the whole context, Mamba struggles to retrieve information from its fixed-size hidden state.
% On the other hand, in Transformers excel as the representation of a token is informed by all preceding tokens.
%Mamba requires approximately double the amount of training data to learn the task every time we increase the sequence length, but also is unable to reach the accuracy threshold for some runs with longer sequences.
%These results show that in tasks requiring access to the whole context, Mamba struggles to retrieve the information its fixed-size hidden state.
%On the other hand, in Transformers the representation of a token is informed by all preceding tokens.
Transformers do not encounter this challenge as the representation of a token is informed by all preceding tokens.

Finally, we discuss how Transformers and Mamba learn to perform in-context retrieval. 
\cref{fig:synthetic_grounding_evolution} illustrates the performance per target token of both models on the synthetic grounding task with sequences of 200 tokens.
Pythia learns the correct target position uniformly. 
On the other hand, Mamba exhibits a different pattern: at the early stages of training it performs adequately in sequences where the target token is located at the end, gradually learns to retrieve the correct token in sequences where the target is at the beginning, and finally, at the end of training learns the task with a target token in between the sequence.

\section{Conclusion}
\paragraph{Implications of Findings} 
In this work, we compare Transformer and SSM-based language model backbones for VLMs.
We show that Mamba consistently outperforms Transformers in tasks where the output depends on a summary of the visual information. 
Transformers, on the other hand, maintain the lead in visual grounding tasks, which we link to their ability to learn more accurately and efficiently to retrieve dense information from the context.
% We consider this as an essential ability because, in some form or another, these models need to generate a response based on the information given. 
% We therefore expect that our findings translate to other tasks and possibly other modalities, where retrieval is a key factor.

% \noindent Regardless, it is important to acknowledge Mamba's memory and computational advantages over Transformers, particularly on long sequence modeling tasks.
\noindent Regardless, Mamba and SSMs, in general, have memory and computational advantages that could be especially critical for tasks that require modeling long sequences, such as high-resolution images, videos, or multimodal documents.
Developing hybrid architectures that integrate an attention-like mechanism into SSMs \citep{dao2024transformers, waleffe2024empirical} is therefore an exciting avenue for future work. 
Such architectures could lead to efficient VLMs that are also able to effectively retrieve relevant information from the context.

\paragraph{Feature or Bug?}
Additionally, we experiment with the effect of placing the instruction before and after the visual input.
While task-aware image encoding provides a marginal performance boost for Mamba on visual grounding, the results fluctuate across other tasks. 
Ultimately, we want multimodal models that can seamlessly encode different modalities without forcing a strict order on how they are presented to the model.
From this perspective, performance differences due to the input structure are a strong signal that the current iteration of VLMs is only partially addressing this issue.

\section*{Limitations}

\paragraph{Data Ablations}
We have not investigated any impact of the data and task distribution. 
We have not covered any ablations regarding how the examples are packed into sequences.
Recent work has shown that this might affect downstream performance in LLMs \citep{zhao2024analysing}. 
Based on our analysis in \cref{sec:vg}, and the conclusions from concurrent work \citep{jelassi2024repeat, merrill2024illusion}, we expect that Transformer and Mamba models might behave differently with different packing strategies.
However, we want to emphasize that both models are trained on the same data thereby ensuring a fair comparison between them, and also the distribution of the data is heavily skewed towards grounding tasks due to the inclusion of the GRIT dataset.

% \paragraph{Lack of multinguality} We would like to emphasize that the pretrained LLMs used in this work are trained using the Pile \cite{gao2020pile} an English-only corpus. 
% Multilinguality is an appealing property for modern VLMs \citep{bai2023qwen, liu2023improved}. 
% However, for reproducibility purposes and a fair comparison between Transformers and Mamba, in this work we have exclusively focused on English benchmarks as well as images that primary reflect western cultures.
% It would be interesting to explore how multilinguality and culture-specific concepts unfold in SSMs and Transformers as a follow-up work.

\section*{Ethics statement}
\paragraph{The Cost of Training Large Scale VLMs} It has been increasingly transparent that the cost of training large-scale models, including VLMs, raises compute barriers \cite{strubell-etal-2019-energy, thompson2020computational, luccioniPower}. 
While patch representations have become the standard approach for encoding images, these representations substantially increase the context window and, consequently, the computational cost of training. 
To improve efficiency, we have employed sequence packing during training, which results to fewer padding tokens within the batch.
Additionally, more sophisticated V\&L connectors that downsample the visual sequence \citep{alayrac2022flamingo, dai2024instructblip, laurenccon2024obelics} can, in principle, accelerate training and inference.
% despite adding more parameters to the connector module as the LLM backbone is presented with significantly fewer tokens.
We leave comparisons of more efficient V\&L connectors in combination with SSMs as future work.
% there is yet an apple-to-apple comparison that shows the effectiveness of this approach.
% Future work concerning the development of a more sample-efficient resampler could pave the way for the new generation of VLMs.

\paragraph{Hallucinations \& Reliability} A widely acknowledged limitation for LLMs and VLMs is the factuality of the generated content \cite{ji2023survey}. 
The impact of this property can vary depending on the downstream task (e.g., answering a question accurately versus creating novel images with text prompts). 
Furthermore, prior work \citep{pantazopoulos2024learning}, has shown that the visual instruction tuning stage imposes a forgetting effect on the safety guardrails of the backbone LLM leading to more vulnerable VLMs.
In this work, we use POPE \citep{li2023evaluating}, a benchmark specifically designed to evaluate object hallucinations in VLMs.
% POPE includes three escalating stages of difficulty, where the input question refers to popular objects, or objects that have strong co-occurrence in datasets used to train VLMs.
% Furthermore, POPE is framed as a binary classification task, meaning that it is straightforward to report zero-shot results since most VLMs have question-answering capabilities.
However, further investigation is needed to evaluate model hallucinations and improve the reliability of VLMs.

\section*{Acknowledgements}
We would like to thank the reviewers for their valuable feedback during the ARR process.
Additionally, this work was supported by the Edinburgh International Data Facility (EIDF) and the Data-Driven Innovation Programme at the University of Edinburgh.
Finally, the authors acknowledge the use of the HWU high-performance computing facility (DMOG) and associated support services in the completion of this work.

\bibliography{latex/ACL2023}
\bibliographystyle{acl_natbib}

\clearpage
\appendix

\section{Datasets}
\label{sec:appedix-dataset}

\begin{table*}[tb]
    \centering
    \small 
    \renewcommand{\arraystretch}{1.3}
    % \begin{adjustbox}{center}
        \begin{tabular}{@{}l l l@{}} 
            \toprule
            \textbf{Task} & \textbf{\# Packed Samples} & \textbf{Dataset} \\ 
            \midrule
            Captioning & 588K & COCO, TexCaps\\
            Chat & 157K & LLaVA-Instruct\\
            Dense Captioning & 467K & RefCOCO, RefCOCO+, RefCOCOg, Visual Genome\\
            Grounded Captioning & 4.2M & GRIT\\
            Image-Text Matching & 8k & VSR\\
            Multiple-Choice VQA & 127K & AI2D, Visual7W\\
            VQA & 352K & \begin{tabular}{@{}l@{}} VQAv2, GQA, OCR-VQA, VG-QA, \\ DocVQA, InfographicVQA\end{tabular}\\
            Visual Grounding & 467K & RefCOCO, RefCOCO+, RefCOCOg, Visual Genome\\
            \midrule
            Total & 6.2M & \\
            \bottomrule
        \end{tabular}
    % \end{adjustbox}
    \caption{Dataset statistics for instruction-tuning. We pack examples from the image into the sequences. }
    \label{tab:instruction-tuning-datasets}
\end{table*}

\begin{table*}[tb]
    \centering
    \small 
    \renewcommand{\arraystretch}{1.3}
    % \begin{adjustbox}{center}
        \begin{tabular}{@{}l l c c@{}} 
            \toprule
            Dataset & Tasks & {\# Images} & {\# Packed Samples}\\ 
            \midrule
             \begin{tabular}{@{}l@{}}AI2D \citep{kembhavi2016diagram}\end{tabular} & Multiple Choice VQA & 3K & 44K\\
             \midrule
             \begin{tabular}{@{}l@{}}A-OKVQA \citep{schwenk2022okvqa}\end{tabular} & Multiple Choice VQA & 16K & 68K\\
             \midrule
            \begin{tabular}{@{}l@{}}COCO \citep{lin2014microsoft}\end{tabular} & Captioning & 113K & 566K\\
            \midrule
            \begin{tabular}{@{}l@{}}DocVQA \citep{mathew2021docvqa}\end{tabular} & VQA & 10K & 20K\\
            \midrule
            \begin{tabular}{@{}l@{}}GQA \citep{hudson2019gqa}\end{tabular}& VQA & 87K & 72K\\
            \midrule
            \begin{tabular}{@{}l@{}}GRIT \citep{peng2023kosmos}\end{tabular} & Grounded Captioning & 4M & 4M\\
            \midrule
            \begin{tabular}{@{}l@{}}InfographicVQA \citep{mathew2022infographicvqa}\end{tabular}& VQA & 4K & 12K\\
            \midrule
            \begin{tabular}{@{}l@{}}LLaVA-Instruct \citep{liu2024visual}\end{tabular} & Chat & 81K & 157K\\
            \midrule
            \begin{tabular}{@{}l@{}}OCR-VQA \citep{mishra2019ocr}\end{tabular}& VQA & 66K & 66K\\
            \midrule
             \begin{tabular}{@{}l@{}}RefCOCO \citep{kazemzadeh2014referitgame}\end{tabular} &  \begin{tabular}{@{}l@{}}Dense Captioning \\ Visual Grounding\end{tabular} & \begin{tabular}{@{}l@{}}16K\\16K\end{tabular} & \begin{tabular}{@{}l@{}}16K\\16K\end{tabular}\\
            \midrule
            \begin{tabular}{@{}l@{}}RefCOCOg \citep{kazemzadeh2014referitgame}\end{tabular} & \begin{tabular}{@{}l@{}}Dense Captioning \\ Visual Grounding\end{tabular} & \begin{tabular}{@{}l@{}}21K\\21K \end{tabular} &  \begin{tabular}{@{}l@{}}21K\\21K\end{tabular}\\
            \midrule
            \begin{tabular}{@{}l@{}}RefCOCO+ \citep{kazemzadeh2014referitgame}\end{tabular}& \begin{tabular}{@{}l@{}}Dense Captioning \\ Visual Grounding\end{tabular} & \begin{tabular}{@{}l@{}}16K\\16K\end{tabular} &  \begin{tabular}{@{}l@{}}16K\\16K\end{tabular}\\
            \midrule
            \begin{tabular}{@{}l@{}}TextCaps \citep{sidorov2020textcaps}\end{tabular} & Captioning & 109K & 21K\\
            \midrule
            \begin{tabular}{@{}l@{}}VQAv2 \citep{goyal2017making}\end{tabular}& VQA & 84K & 82K\\
            \midrule
            \begin{tabular}{@{}l@{}}VSR \citep{liu2023visual}\end{tabular}& Image Text Matching & 5k & 8K\\
            \midrule
            \begin{tabular}{@{}l@{}}Visual Genome \\ \citep{krishna2017visual}\end{tabular}& \begin{tabular}{@{}l@{}}Dense Captioning \\ Visual Grounding \\ VQA \end{tabular} &  \begin{tabular}{@{}l@{}}411K\\411K\\184K\end{tabular} &  \begin{tabular}{@{}l@{}}105K\\105K\\97K\end{tabular}\\
            \midrule
            \begin{tabular}{@{}l@{}}Visual7W \citep{zhu2016visual7w}\end{tabular} & Multiple Choice VQA & 27K & 255K\\
            \midrule
            Total &  &  & 5.7M\\
            \bottomrule
        \end{tabular}
    % \end{adjustbox}
    \caption{Dataset statistics for instruction-tuning. We pack image-text examples from the same dataset into the same sequence.}
    \label{tab:data-mixture}
\end{table*}

\subsection{Data Mixture}

 \cref{tab:instruction-tuning-datasets} shows the datasets used for instruction tuning.
 \cref{tab:data-mixture} shows a detailed breakdown regarding the number of examples for each task.

\begin{figure}[tb]
    \centering
    \includegraphics[width=\linewidth]{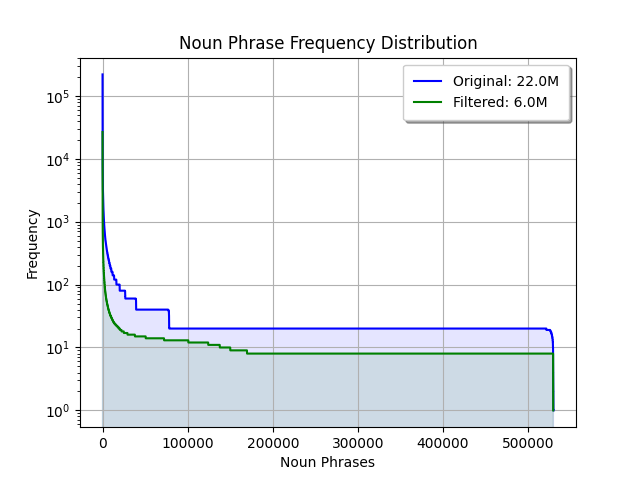}
    \caption{Filtered and unfiltered distribution of noun phrases in GRIT. By applying minimal filtering, we can reduce the dataset size while at the same time maintain object coverage.}
    \label{fig:grit-filter}
\end{figure}

\begin{table*}[tb]
    \centering
    \small
    \renewcommand{\arraystretch}{1.3}
    % \begin{adjustbox}{center}
        \begin{tabular}{@{}l l@{}} 
            \toprule
            Task & Instruction\\
            \midrule
            Captioning & Provide a one-sentence caption for the provided image\\
            Dense Captioning & Provide a short description of the region\\
            Grounded Captioning & Provide a one-sentence caption for the image and mention each entity.\\
            Image Text Match & Determine if the image matches the description\\
            Multiple Choice VQA & Answer with the option's letter from the given choices directly\\
            Visual Grounding & Locate the region that is described by\\
            Visual Question Answering & Answer the question using a single word or phrase \\
            \bottomrule
        \end{tabular}
    % \end{adjustbox}
    \caption{Instructions for all tasks.}
    \label{tab:instruction}
\end{table*}

\begin{table*}[tb]
     \small
     \centering
     \renewcommand{\arraystretch}{1.3}
     \begin{tabular}{@{}l l l@{}}
     \toprule
     \textbf{DC} & \raisebox{-0.5\height}{\includegraphics[width=0.3\textwidth]{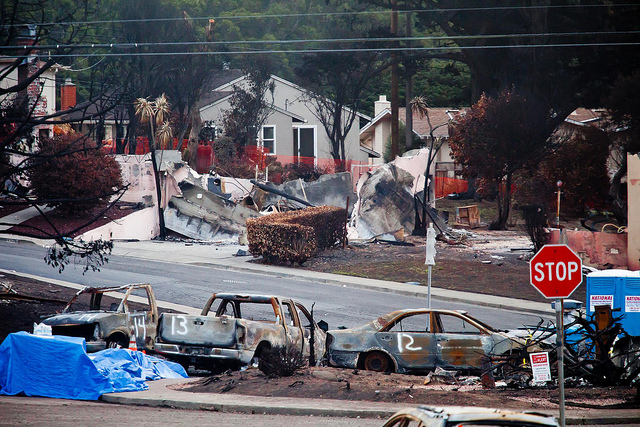}} &  \begin{tabular}{@{}l@{}}Provide a short description of the region\\$[0.50, 0.72, 0.87, 0.89]$\\\colorbox{lightmagenta}{A rusted junk car with a white R painted on the door}\\$[0.24, 0.69, 0.51, 0.87]$\\\colorbox{lightmagenta}{A rusted truck with '13' spray painted on it.}\\$[0.23, 0.68, 0.50, 0.88]$\\\colorbox{lightmagenta}{The pick-up marked 13}\end{tabular}\\
     % \textbf{VG} & \raisebox{-0.5\height}{\includegraphics[width=0.3\textwidth]{figures/2322427.jpg}} & \begin{tabular}{@{}l@{}}Provide a short description of the region\\ $[0.39, 0.244, 0.41 , 0.27]$\\\colorbox{lightmagenta}{Dog's eye is black}\\Black collar on dog\\\colorbox{lightmagenta}{[0.45 , 0.24, 0.52 , 0.42]}\\Dog's tail pointing upwards\\\colorbox{green!20}{[0.48 , 0.00, 0.56, 0.25]}\\Dog's paw off the ground\\\colorbox{green!20}{[0.42 , 0.54, 0.50 , 0.67]}\end{tabular}\\
     \midrule
     \textbf{M-VQA} & \raisebox{-0.5\height}{\includegraphics[width=0.3\textwidth]{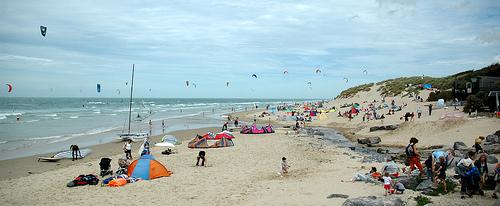}} & \begin{tabular}{@{}l@{}}Answer with the option's letter from the given choices directly\\Question: What color is the closest tent?\\A: Orange and blue. B: White. C: Black. D: Purple.\\Answer: \colorbox{lightmagenta}{A}\\Question: Why is the sand darker at the edge of the ocean?\\A: It is dirty. B: It is wet. C: It's dark out. D: There's a shadow on it.\\Answer: \colorbox{lightmagenta}{B}\\Question: When was this picture taken?\\A: During the night. B: In daytime. C: At dawn. D: At dusk.\\Answer: \colorbox{lightmagenta}{B}\end{tabular}\\
     \midrule
     \textbf{VG} & \raisebox{-0.5\height}{\includegraphics[width=0.3\textwidth]{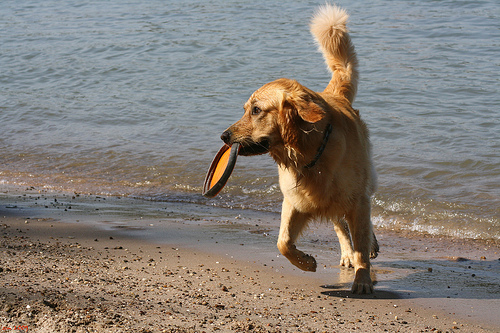}} & \begin{tabular}{@{}l@{}}Locate the region that is described by\\Dog's eye is black\\\colorbox{lightmagenta}{[0.39, 0.24, 0.41 , 0.27]}\\Black collar on dog\\\colorbox{lightmagenta}{[0.45 , 0.24, 0.52 , 0.42]}\\Dog's tail pointing upwards\\\colorbox{lightmagenta}{[0.48 , 0.00, 0.56, 0.25]}\\Dog's paw off the ground\\\colorbox{lightmagenta}{[0.42 , 0.54, 0.50 , 0.67]}\end{tabular}\\
      \midrule
     \textbf{VQA} & \raisebox{-0.5\height}{\includegraphics[width=0.3\textwidth]{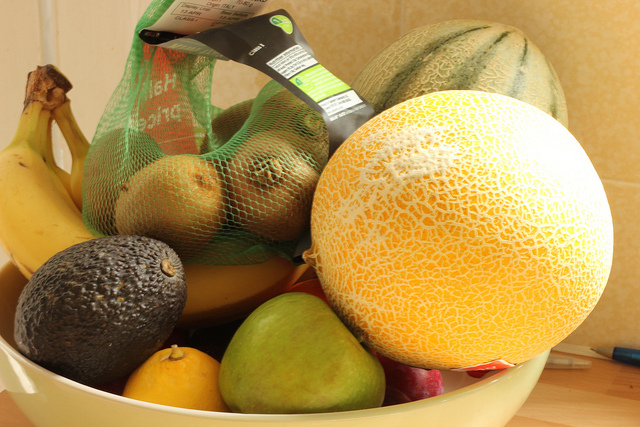}} & \begin{tabular}{@{}l@{}}Answer the question using a single word or phrase\\Question: Are all the items in the bowl fruits?\\Answer: \colorbox{lightmagenta}{Yes}\\Question: What is the light green item?\\Answer: \colorbox{lightmagenta}{Apple}\\Question: What is the biggest fruit here\\Answer: \colorbox{lightmagenta}{Cantaloupe}\end{tabular}\\
      \bottomrule
      \end{tabular}
      \caption{Illustration of packing examples for each task. \colorbox{lightmagenta}{Text} are the targets for the model for each example.}
      \label{tab:inputs-outputs}
  \end{table*}

\paragraph{Filtering Visual Genome} We follow OFA \citep{wang2022ofa} by preprocessing region descriptions. 
Specifically, we use only image-region pairs from Visual Genome where the area of the region is smaller than $16,384$ pixels to encourage more fine-grained alignments between vision and language.

\paragraph{Filtering GRIT} The original version of GRIT \citep{peng2023kosmos} contains 20.5M image-grounded caption pairs. 
Simply including this benchmark bears the risk of task imbalance, and therefore overfitting on a given task at very early stages of training \citep{raffel2020exploring}.
Furthermore, previous work \citep{abbas2023semdedup} has shown that semantic deduplication of a large-scale corpus from the web, can significantly reduce the training cost while at the same time maintain performance.

Therefore, to accelerate training without sacrificing diversity we filter GRIT by trying to maximize the number of concepts in the corpus.
An easy approach would be to rank the image-text pairs using CLIPScore \citep{hessel2021clipscore} and then select the top-N images as the filtered corpus. 
However, this approach may result in selecting images of the most frequent concepts and thus do not expose the model to a variety of examples.
For this purpose, we filter the dataset using the noun phrases from each caption.
First, we discard all images with width or height less than 100 pixels.
With regards to the text descriptions, we begin by removing any articles from the noun phrase and then counting all phrases for each image.
Next, starting from the rarest noun phrases: 1) if the frequency is between a \texttt{min} and a \texttt{max} threshold we add all images to our filtered corpus that contain the phrase in their caption, 2) else if the frequency is higher than the maximum threshold we randomly select \texttt{max} images.
As shown in \cref{fig:grit-filter}, by setting \texttt{min} = 3, and \texttt{max} = 8, we can obtain a smaller corpus that covers all noun phrases.

\paragraph{Filtering OCRVQA} We filter out images with a width or height of less than 350 pixels.
Additionally, we have observed that some image URLs contain blank images (i.e., images with only a single color). 
We performed rudimentary filtering by removing all images that have more than 85\% pixels from the same color. 
Finally, we removed all questions associated with the category of the book (e.g, ``Is this a sociopolitical book?'') as we identified from manual inspection that answering this question based solely from the cover of the book can be particularly challenging.

\paragraph{Multiple Choice VQA} For the multiple choice VQA datasets used in instruction tuning (e.g, AI2D \citep{kembhavi2016diagram}, Visual7W \citep{zhu2016visual7w}, and A-OKVQA \citep{schwenk2022okvqa}), we have augmented the training data by assigning the correct option to all possible character options.
For example, if the question has four candidate answers (A, B, C, D) and the correct answer is A, we created four data points from this question alone by rotating the labels clockwise until the correct answer is in all positions.

\subsection{Response Formatting}
\cref{tab:instruction} shows the instructions used in our models. 
Across all experiments, including the first training stage, we mask the instruction prompts and predict only the response. 
The full sequence given to the model has the following format: \#\#$p_{11}, p_{12}, \dots, p_{1N} \&\& \dots, p_{N1}, p_{N2}, \dots, p_{NN}$\#\# \texttt{<Task Instruction>} \texttt{<Prompt>} \texttt{<Response>}, where the tokens $p_{ij}$ are the embeddings for each patch.

\paragraph{Representing Coordinates in Images} We follow previous VLMs that choose to represent coordinates in images using decimal values \citep{chen2023shikra, bai2023qwen}.
Other works \citep{chen2022pixseq, wang2022ofa, yang2022unitab, peng2023kosmos} introduce special tokens that represent image coordinates in a discrete format. 
This approach increases the size of the model by adding extra rows to the embedding matrix corresponding to the new special tokens.
Furthermore, Shikra \citep{chen2023shikra} has shown preliminary results on the benefits of decimal representation.
While there is yet a comprehensive comparison, we believe that the advantage of the decimal representation is due to the fact that the LLM has often already trained embeddings for the decimal tokens, i.e, the model roughly knows what ``0.5'' refers to and therefore starts from an advantageous point during the visual instruction tuning stage.
However, decimal representation introduces longer sequences which prolongs training and inference.
Future work could further explore this trade-off.

\subsection{Dataset Packing}
A significant component during our model development is how we pack the examples into sequences in a meaningful way.
% Packing examples into longer benefits has been identified as a key factor in accelerating training by minimizing unnecessary computation due to padding tokens in a batch \citep{krell2021efficient}.
The benefits of this dataset packing are two-fold: 1) we ensure efficiency in training by minimizing unnecessary computations due to the padding tokens in a batch \citep{krell2021efficient}, and 2) by packing examples we facilitate chat capabilities of our models to some degree.
In this work, we pack examples from the same image into a sequence of input-output pairs.
As already mentioned, we apply packing for all (multiple-choice) VQA, Visual Grounding, and Dense Captioning examples. 
We refrain from packing captioning examples because the target captions can be repetitive, therefore the model may rely on previous captions without paying attention to the image. 
We aimed for a maximum sequence length of 1024 tokens including the patch embeddings and the special image tokens.
For this purpose packed examples from VQAv2 \citep{goyal2017making} are limited to 20 qa pairs. 
Similarly, we limit the number of qa-pairs to 10 and 5 for the telling, and pointing task in Visual 7W \citep{zhu2016visual7w}.
Finally, for all tasks in Visual Genome \citep{krishna2017visual}, all packed examples are limited to 10 input-output responses.

\section{Training Details}\label{appendix:training-details}

\begin{table}[tb]
    \centering
    % \scriptsize
    \small
    \renewcommand{\arraystretch}{1.3}
    % \begin{adjustbox}{center}
        \begin{tabular}{@{}l c c l@{}} 
            \toprule
            Hyperparameter & Pretraining & Instruction Tuning\\
            \midrule
            global batch size & 256 & 128\\
            lr & 1e-3 & 2e-5\\
            lr schedule & \multicolumn{2}{c}{cosine decay} \\
            lr warmup & 0.03\\
            number of epochs & 5 & 2\\
            optimizer & \multicolumn{2}{c}{AdamW}\\
            DeepSpeed Stage & 2 & 3\\
            \bottomrule
        \end{tabular}
    % \end{adjustbox}
    \caption{Hyperparameters during both training stages. The same hyperparameters are used for Pythia-VL, and Mamba-VL across all three different scales.}
    \label{tab:training-hyperparams}
\end{table}

\begin{figure}[tb]
    \centering
    \includegraphics[width=0.45\textwidth]{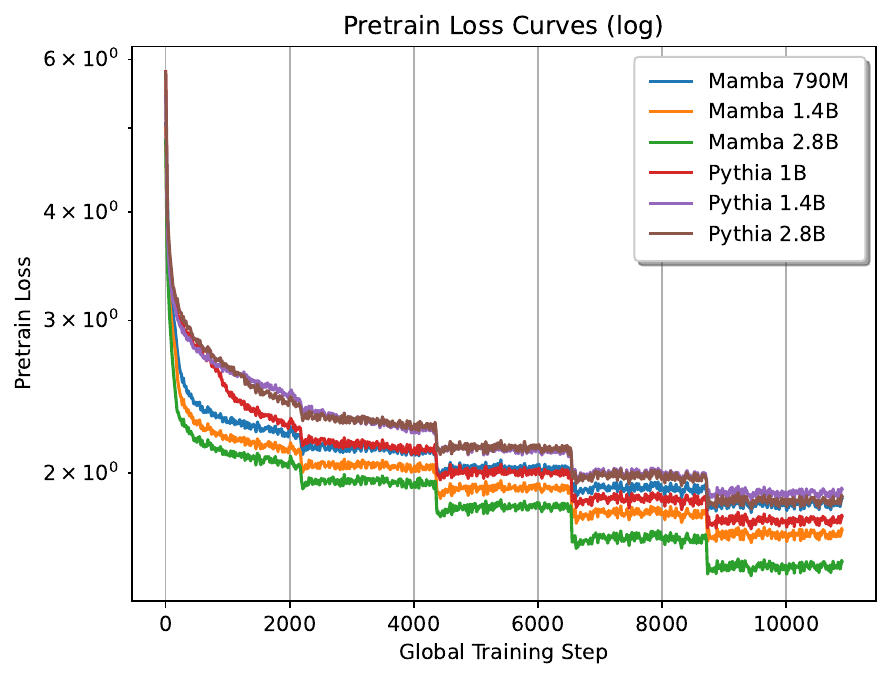}
    \caption{Loss curves for all models during pretraining.}
    \label{fig:pretrain-loss-curves}
\end{figure}

% \begin{figure}[ht!]
%     \centering
%     \includegraphics[width=0.45\textwidth]{figures/fine_tune_train_grad_curves_s100_axes.pdf}
%     \caption{Gradient norm during finetuning.}
%     \label{fig:grad-norm}
% \end{figure}

% \begin{figure}[ht!]
%     \centering
%     \includegraphics[width=0.45\textwidth]{figures/fine_tune_train_loss_curves_s100_axes.pdf}
%     \caption{Gradient norm during finetuning.}
%     \label{fig:grad-norm}
% \end{figure}

\begin{figure*}[tb]
\minipage{0.33\textwidth}
  \includegraphics[width=\linewidth]{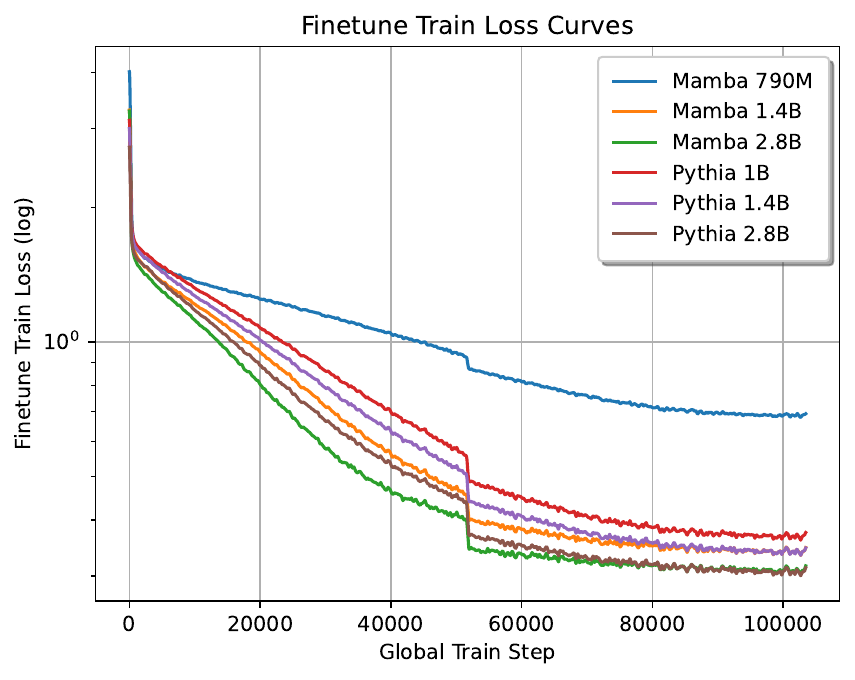}
  \subcaption{Gradient Norm.}\label{fig:awesome_image1}
\endminipage\hfill
\minipage{0.33\textwidth}
  \includegraphics[width=\linewidth]{figures/fine_tune_train_loss_curves_s100_axes_tight.pdf}
  \subcaption{Train Loss.}
\endminipage\hfill
\minipage{0.33\textwidth}%
  \includegraphics[width=\linewidth]{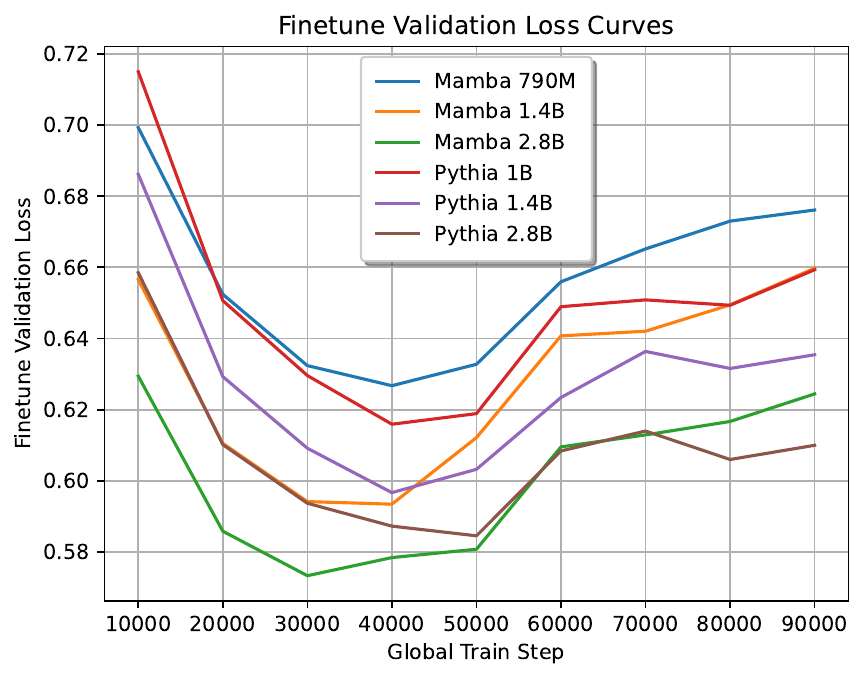}
  \subcaption{Validation Loss.}
  % \caption{A really Awesome Image}\label{fig:awesome_image3}
\endminipage
    \caption{Performance curves for all models during finetuning.}
    \label{fig:finetuning}
\end{figure*}

\paragraph{Training Hyperparameters} We use the same hyperparameters as LLaVA-1.5 for pretraining / instruction tuning. We decided to increase the number of epochs in the pretraining stage as we observed significant performance differences after zero-shot evaluation on COCO captioning with preliminary experiments. 
Additionally, we obtain visual features from the last layer of the EVA model. 
We have not conducted any ablations considering the layer from which to obtain visual representations. 
All experiments were conducted using 4x A100 (40GB / 80GB) or 2x H100 GPUs.
For the small models ($\le 1.4$B) we set the maximum sequence length to 1024. 
For the larger models (2.8B) we set the maximum sequence length to 800 to maintain a large batch size during training. 
Note that this results in a small loss of within-sequence examples.
We pretrain each model for 10k steps. 
We train each model for 100k during the instruction tuning phase, where we evaluate each checkpoint after 10k steps.
However, we found that the latest checkpoint resulted in greater performance across both models, despite the higher validation loss.

\paragraph{Model Training Strategy} We employ a `mixed batches' approach, where a batch contains examples from any instruction tuning task. However, we have not used any form of custom sampling e.g adjusting the sampling weight based on the size of the dataset \citep{raffel2020exploring}.
Additionally, we note that the target length can vary significantly per task, for example the correct response to multiple choice VQA is a single token (e.g, the character from the given options), while for captioning examples the target sequence is longer.
Therefore, similar to previous work \citep{ustun2024aya, pantazopoulos2023multitask}, we normalize the cross-entropy loss over the target tokens per sequence first and then average over all the sequences in the batch to weigh all samples equally during finetuning.

\paragraph{Training Logs}
All training logs regarding pretraining and instruction turing are available \href{https://wandb.ai/gpantaz/vl_mamba?nw=nwusergpantaz}{here}. 
We also provide here the training curves for the pretraining (\cref{fig:pretrain-loss-curves}) and instruction tuning for all models (\cref{fig:finetuning}).
In all of our cases the latest model achieved the best performance despite the trend in the validation loss.

\paragraph{Finetuning on Downstream Tasks} In \cref{sec:results} we also report the results of Pythia-VL and Mamba-VL with 1.4B parameters on VQAv2 and RefCOCOg.
For this purpose, we apply a small grid search for each task by using three values for the learning rate ($1e^{-5}$, $5e^{-5}$, $1e^{-4}$) and a batch size of 64. 
We finetune each checkpoint from the instruction tuning stage for 1 and 3 epochs on VQAv2 and RefCOCOg, respectively, by keeping the examples packed into larger sequences.
To increase the resolution of images we simply interpolated the positional embeddings of the vision encoder.
We did not scale the rotary embeddings of Pythia, as even in the highest resolution images (560) the sequence length does not exceed the maximum sequence length of the pretrained language model.
The models with higher resolution (560) are not using the checkpoint from the finetuning of the previous lower resolution (448).
We report the best performing model on the validation split of VQAv2 and test split of RefCOCOg.

\section{Experiments}\label{appendix-experiments}

\subsection{Benchmarks \& Metrics}
\cref{tab:benchmark_metrics} shows the benchmarks used for our evaluation with their respective metrics.

\begin{table}[tb!]
    \centering
    \small
    % \small
    \renewcommand{\arraystretch}{1.3}
    % \begin{adjustbox}{center}
    \resizebox{\linewidth}{!}{
        \begin{tabular}{@{}lll@{}} 
            \toprule
            Benchmark & Zero-shot & Metrics\\
            \midrule
            COCO & \xmark & CIDEr (C), BLEU-4 (B4), METEOR (M),\\
            & &  ROUGE (S), Spice (S)\\
            NoCaps & \cmark & CIDEr\\
            \midrule
            VQAv2 & \xmark & VQAv2 score\\
            GQA & \xmark & Accuracy\\
            Visual7W (T) & \xmark & Accuracy (Multiple Choice)\\
            \midrule
            VSR & \xmark & Accuracy \\
            POPE & \cmark & Accuracy \\
            \midrule
            RefCOCO /g/+ & \cmark & Accuracy@IoU$\ge0.5$\\
            \midrule
            TextCaps & \cmark & CIDEr\\
            TextVQA & \cmark & Accuracy \\
            AI2D & \cmark & Accuracy (Multiple Choice)\\
            \bottomrule
        \end{tabular}
    }
    % \end{adjustbox}
    \caption{Evaluation metrics for each benchmark.}
    \label{tab:benchmark_metrics}
\end{table}

\begin{table}[tb!]
    \centering
    \small
    % \small
    \renewcommand{\arraystretch}{1.3}
    % \begin{adjustbox}{center}
        \begin{tabular}{@{}l ccccc@{}} 
            \toprule
            Model & C & B4 & M & R & S\\
            \midrule
            Mamba-CLIP & 79.9 & 21.8 & 22.2 & 47.9 & 16.5\\
            Mamba-EVA02 & \textbf{87.1} & \textbf{23.7} & \textbf{23.2} & \textbf{48.8} & \textbf{17.9}\\
            \bottomrule
        \end{tabular}
    % \end{adjustbox}
    \caption{Performance of Mamba-790M on COCO test after the first training stage using similar sized CLIP and EVA02 models. Across the board, Mamba achieves greater performance when paired with EVA.}
    \label{tab:evavsclip}
\end{table}

\subsection{Comparison between EVA-02 and CLIP}
We evaluate a Mamba-790M checkpoint after the first training stage using EVA-02 Large 336px/14 \citep{fang2023eva} and CLIP-Large 336px/14 \citep{radford2021learning}. 
We use the same training parameters across both runs. 
\cref{tab:evavsclip} illustrates the results on COCO without any fine-tuning.
We observe that using visual representations from EVA leads to greater performance.

\subsection{Task-agnostic Visual Encoding}\label{appendix:task-aware}
We provide the full results showcasing a comparison between task-agnostic and task-aware visual encoding, where the task identity is known to the model before encoding images. 
\cref{tab:task-agnostic-full} illustrates the performance for each model with and without task-agnostic visual encoding for all held-in benchmarks.
We would like to highlight that a similar comparison has been conducted for Transformer-based VLMs in InstructBLIP \citep{dai2024instructblip}, showcasing that the task-aware visual encoding is beneficial in held-in as well as held-out benchmarks.
However, InstructBLIP opts for a specific architectural choice, where the task-aware encoding is conducted at the connector module between the LLM and the vision encoder.
The connector (i.e the QFormer), is creating a multimodal prompt that is then prepended to the instruction at the input of the LLM.
This means that in practice the LLM sees first the visual prompt and then the instruction.
This architectural choice might justify the need for more suitable and versatile multimodal fusion architectures.

\begin{table*}[tb]
    \centering
    % \scriptsize
    \tiny
    \renewcommand{\arraystretch}{1.2}
    % \begin{adjustbox}{center}
    \addtolength{\tabcolsep}{-0.3em}
        \begin{tabular}{@{}ll |c| ccc| c| cc cc cc| cc@{}} 
            \toprule
            & & \textbf{Image Captioning} & \multicolumn{3}{c|}{\textbf{General VQA}} &  \textbf{Misc} & \multicolumn{6}{c|}{\textbf{Visual Grounding}} & \multicolumn{2}{c}{\textbf{Reading Comprehension}}\\
            Model & Task & COCO &  VQAv2 & GQA &  V7W & VSR &  \multicolumn{2}{c}{RefCOCO} & \multicolumn{2}{c}{RefCOCO+} & RefCOCOg & V7W (P) & TextCaps & AI2D\\
            & Agnostic & test & val & test-dev & test-T & test & testA & test B & testA & testB & test-P & test & val & test\\
            \midrule
            Pythia-VL & \cmark & 134.06 & 73.57 & 57.05 & 83.06 & 77.72 & 82.43 & 68.39 & 72.35 & 55.16 & 72.56 & 86.13 & 94.60 & 79.27\\
            Pythia-VL & \xmark & 133.87 & \textbf{73.15} & 58.12 & 79.30 & 76.94 & 82.78 & 68.89 & 71.74 & 54.44 & 73.76 & 85.41 & 95.03 &  79.83\\
            \midrule
            Mamba-VL & \cmark & 134.76 & 74.46 & 58.44 & 83.78 & 80.18 & 76.60 & 63.48 & 68.40 & 52.11 & 68.82 & 80.18 & 98.68 & 80.20\\
            Mamba-VL & \xmark & 135.45 & \textbf{74.58} & 58.32 & 83.19 & 79.54 & 77.77	& 65.35 & 68.25	& 51.79 & 70.01 & 77.04 & 100.2 & 80.86\\
            \midrule
            \multicolumn{15}{c}{\textbf{Relative Performance Gain Per Task}} \\
            \midrule
            Pythia-VL & - & -0.14 & -0.57 & +1.88 & -4.53 & -1.00 & +0.42 & +0.73 & -0.84 & -1.31 & +1.10 & -0.84 & -0.45 & -0.70\\
            Mamba-VL & - & +0.51 & +0.16 & -0.21 & -0.70 & -0.80 & +1.53 & +2.95 & -0.22 & -0.61 & +1.73 & +3.92 & +1.54 & -0.82\\
            % \midrule
            %  \multicolumn{15}{c}{\textbf{Mean Relative Performance Gain Per Task}} \\
            %  \midrule
            %  Pythia-VL & - & -0.14 & \multicolumn{3}{c|}{-1.07} & -1.00 & \multicolumn{6}{c|}{-0.12} & \multicolumn{2}{c}{-0.12}\\
            %  Mamba-VL & - & +0.51 & \multicolumn{3}{c|}{-0.25} & -0.80 & \multicolumn{6}{c|}{+0.24} & \multicolumn{2}{c}{+1.18}\\
            \bottomrule
        \end{tabular}
    % \end{adjustbox}
    \caption{Comparison of Pythia-VL \& Mamba-VL with task-agnostic and task-aware visual encoding.}
    \label{tab:task-agnostic-full}
\end{table*}

\subsection{Synthetic Grounding}\label{appendix:synth_vg}
For the task of synthetic grounding, we create sequences of varying lengths (50/100/200).
For each sequence, we created in total 1M training examples and evaluated each model on 100k held-out samples. 
To eliminate any biases regarding the distribution of the targets, we equally distributed the target token evenly within the sequence. 
For example, for sequences with 100 tokens, 1\% of the training examples (1000) have the 1st token as target. 
All models are trained using a global batch size of 64 for 78K steps. 
We evaluated every model after 1\% of training steps to capture precisely the timestep where each model learns the task.

\paragraph{Relation to Induction Heads} Our task is closely related to the Induction Heads \citep{olsson2022context}, which requires models to perform associative recall by retrieving relevant information from the memory.
More specifically, if the model has already observed the pattern \texttt{AB} in a sequence of tokens, then it should be able to infer that \texttt{A} is followed by \texttt{B} some time within the same sequence.

The results of Mamba \citep{gu2023mamba} on Induction Heads show that a two-layer model trained on short sequences maintains high performance across varying sequence lengths compared to other SSMs and Transformer recipes.
A key difference between this setup and how we framed our synthetic grounding task is that in the Induction Heads benchmark there exists a single special token in a sequence and the model always needs to predict the follow-up token (e.g, Input: a b c d e $ \vdash$ f g h i $\dots$ x y z $\vdash$, Output: f \citep{fu2023hungry}). 
On the other hand, in our task every token in the sequence is a ``special token'', and the model needs to be able to recall every element in order but also perform a two-hop reasoning between two embedding spaces.

\subsubsection{Prefix Variation}\label{sec:appendix-prefix}
\begin{figure*}[tb]
\minipage{0.33\textwidth}
  \includegraphics[width=\linewidth]{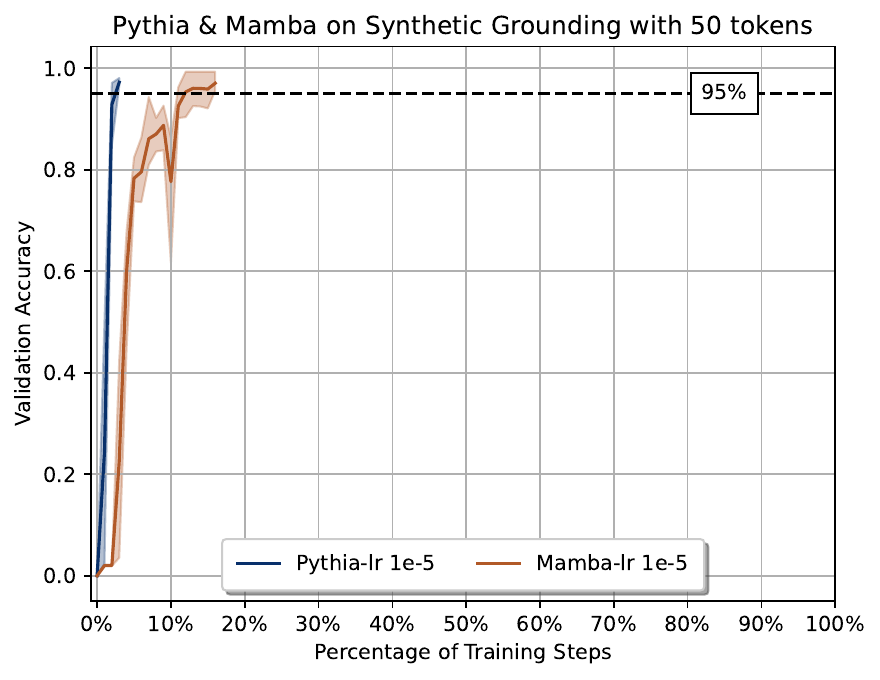}
  \subcaption{Sequence length = 50.}
\endminipage\hfill
\minipage{0.33\textwidth}
  \includegraphics[width=\linewidth]{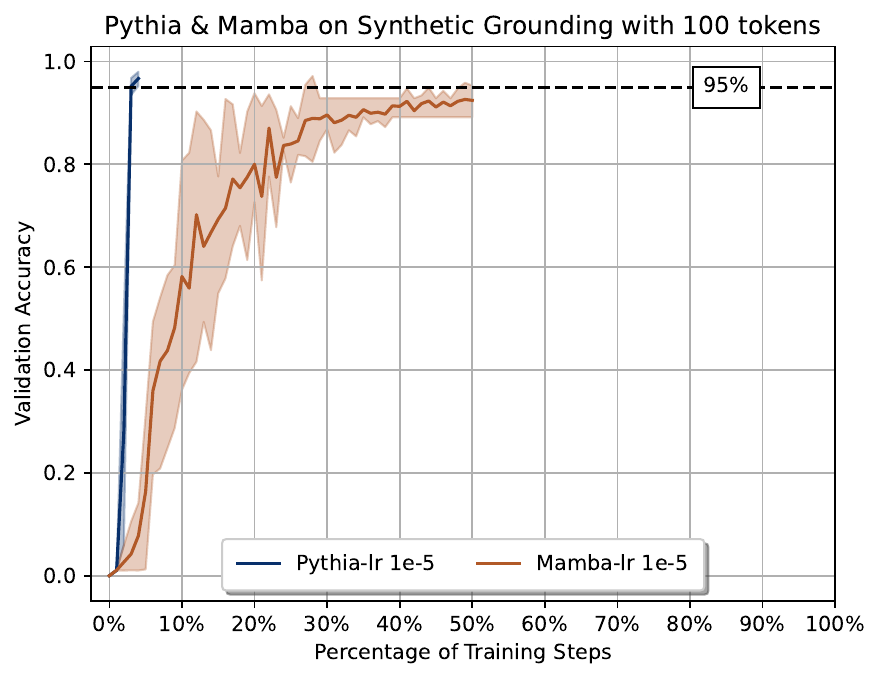}
  \subcaption{Sequence length = 100.}
\endminipage\hfill
\minipage{0.33\textwidth}%
  \includegraphics[width=\linewidth]{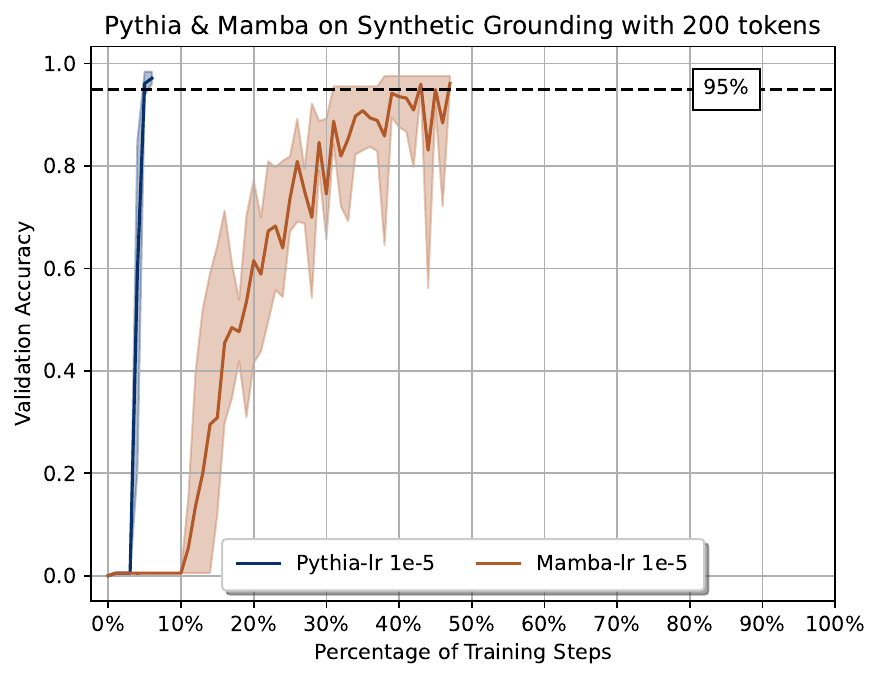}
  \subcaption{Sequence length = 200.}
\endminipage
    \caption{Performance curves for Pythia-1.4B and Mamba-1.4B variants on the synthetic grounding task with varying sequence length and the prefix modification. }
    \label{fig:prefix_synthetic_grounding_results}
\end{figure*}

\begin{figure}[tb]
    \centering
    \includegraphics[width=0.8\linewidth]{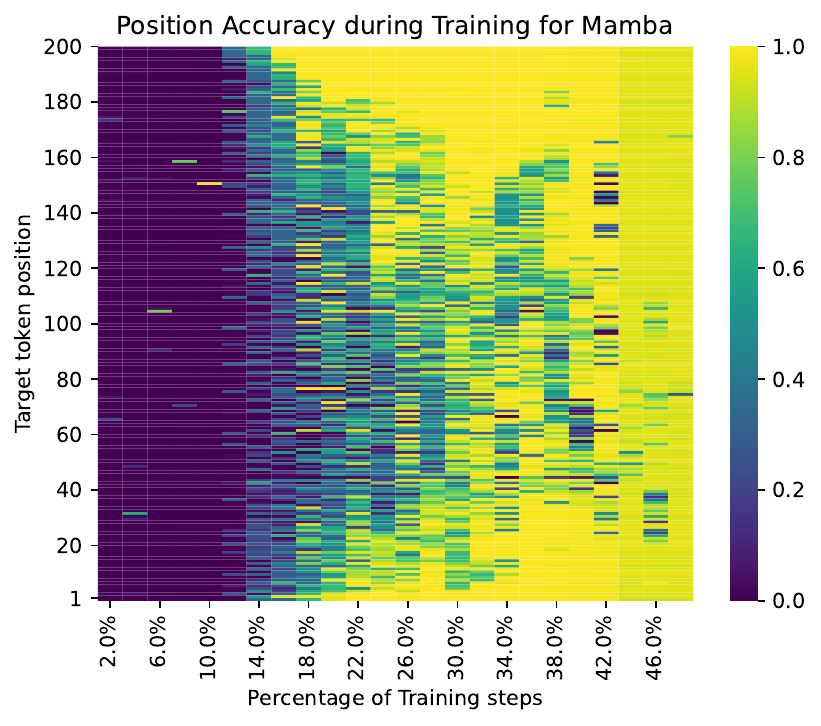}
    \caption{Accuracy per position on the held-out set during training on sequences of 200 tokens the prefix synthetic grounding task with the prefix modification.}
    \label{fig:per_position_prefix}
\end{figure}

Additionally, motivated by the improvements of the task-aware encoding on visual grounding, we experiment with a prefix variant of our synthetic task.
The key difference is that the query precedes the input sequence, and therefore, Mamba has direct access to the required information from the beginning.
We experiment with the same sequence lengths for Pythia and Mamba. 
\cref{fig:per_position_prefix} illustrates the performance of both models. 
Compared to the suffix variant (\cref{fig:synthetic_grounding_evolution}) we can see that Mamba learns the task significantly faster.
For example, in the suffix version of the task and for sequences of 200 tokens, Mamba is not able to reach 95\% accuracy in the training window. 
On the other hand, in the prefix setting and for the same sequence length, we observe that Mamba learns the task within the first half of the training.
Nevertheless, even on this setup, Pythia is more efficient as it learns the task within only 10\% of the training steps.

\end{document}